\documentclass[]{fairmeta}
\usepackage{graphicx}
\usepackage{tabularx}

\usepackage{colortbl}
\usepackage{tcolorbox}
\usepackage{inconsolata}
\usepackage{amsmath}
\usepackage{arydshln}
\usepackage{enumitem}
\usepackage{listings}
\usepackage{fancyvrb}
\usepackage{framed}
\usepackage{float}

\definecolor{verylightgray}{rgb}{0.9,0.9,0.9}

\newcolumntype{P}[1]{>{\centering\arraybackslash}p{#1}}
\DeclareMathOperator*{\argmax}{\arg\!\max}

\title{A Systematic Examination of Preference Learning through the Lens of Instruction-Following}

\author[1,2]{Joongwon Kim}
\author[1]{Anirudh Goyal}
\author[1]{Aston Zhang}
\author[1]{Bo Xiong}
\author[1]{Rui Hou}
\author[1]{Melanie Kambadur}
\author[1]{Dhruv Mahajan}
\author[2]{Hannaneh Hajishirzi}
\author[1]{Liang Tan}

\affiliation[1]{Llama Team, AI @ Meta}
\affiliation[2]{University of Washington}

\abstract{
Preference learning is a widely adopted post-training technique that aligns large language models (LLMs) to human preferences and improves specific downstream task capabilities.
In this work we systematically investigate how specific attributes of preference datasets affect the alignment and downstream performance of LLMs in instruction-following tasks.
We use a novel synthetic data generation pipeline to generate 48,000 unique instruction-following prompts with combinations of 23 verifiable constraints that enable fine-grained and automated quality assessments of model responses.
With our synthetic prompts, we use two preference dataset curation methods -- rejection sampling (RS) and Monte Carlo Tree Search (MCTS) -- to obtain pairs of (chosen, rejected) responses.
Then, we perform experiments investigating the effects of (1) the presence of shared prefixes between the chosen and rejected responses, (2) the contrast and quality of the chosen, rejected responses and (3) the complexity of the training prompts.
Our experiments reveal that shared prefixes in preference pairs, as generated by MCTS, provide marginal but consistent improvements and greater stability across challenging training configurations.
High-contrast preference pairs generally outperform low-contrast pairs; however, combining both often yields the best performance by balancing diversity and learning efficiency.
Additionally, training on prompts of moderate difficulty leads to better generalization across tasks, even for more complex evaluation scenarios, compared to overly challenging prompts.
Our findings provide actionable insights into optimizing preference data curation for instruction-following tasks, offering a scalable and effective framework for enhancing LLM training and alignment.
}

\date{\today}
\correspondence{Joongwon Kim at \email{jwonkim@meta.com}}

\begin{document}

\maketitle

\section{Introduction}
\label{section:intro}
Aligning large language models (LLMs) with human preferences has remained a persistent challenge despite their recent success, particularly for tasks that involve generating nuanced, instruction-following responses.
To address this bottleneck, \textbf{preference learning} has emerged as a vital technique applied in the final stages of LLM post-training~\citep{DBLP:conf/nips/StiennonO0ZLVRA20, DBLP:conf/nips/Ouyang0JAWMZASR22, DBLP:journals/corr/abs-2204-05862}.
Preference learning refines the ability of LLMs to align with human expectations by fine-tuning them on pairs of (chosen, rejected) responses.
Recent successes of using preference learning to develop frontier language models~\citep{DBLP:journals/corr/abs-2303-08774, DBLP:journals/corr/abs-2312-11805, Anthropic2024, DBLP:journals/corr/abs-2407-21783} have led to automated methods for curating preference pairs~\citep{DBLP:journals/corr/abs-2404-19733, DBLP:journals/corr/abs-2405-00451, DBLP:journals/corr/abs-2312-16682, DBLP:conf/naacl/KhakiLMYR24}.
While these techniques yield synthetic preference pairs that significantly improve model capabilities in closed-ended tasks, it is yet unclear which attributes of the preference pairs contribute to the improved alignment and downstream capabilities.

Existing research in preference learning has largely focused on the mechanics of optimization methods, such as Direct Preference Optimization (DPO,~\cite{DBLP:conf/nips/RafailovSMMEF23}) and Proximal Policy Optimization (PPO,~\cite{DBLP:journals/corr/SchulmanWDRK17}).
While these methods are critical for updating model weights based on preference data, they operate with limited insight into how the structure, quality, and complexity of the preference datasets themselves affect outcomes~\citep{DBLP:journals/corr/abs-2406-09279, DBLP:journals/corr/abs-2410-15595}.
For example, questions remain about whether shared prefixes in paired responses improve learning, whether training on high-contrast pairs is always optimal, or how the difficulty of training prompts impacts generalization.
Without a systematic investigation of these factors, designing effective preference datasets is largely heuristic and suboptimal.

In this work we seek to fill this gap by conducting a systematic investigation of how various attributes of automatically-curated preference datasets affect model performance.
We approach this problem from the viewpoint of \textbf{instruction-following}~\citep{DBLP:journals/corr/abs-2311-07911, DBLP:journals/corr/abs-2407-03978, DBLP:journals/corr/abs-2408-01122}, which are ideal for such analysis due to their complexity and capacity for integrating multiple constraints, giving us finer control over the data compared to other domains with binary correctness such as mathematics or tool use~\citep{DBLP:conf/nips/HendrycksBKABTS21, DBLP:conf/iclr/MialonF0LS24}.
Furthermore, we focus on \textit{verifiable} constraints that can be assessed by code, which allows us to enable precision by deterministically evaluating the quality of any response with respect to the constraints, and scalability by requiring much less compute than open-ended constraints.

We first define an ontology of 23 verifiable constraints spanning diverse requirements such as adherence to specific structural, stylistic, or formatting requirements, yet \textit{distinct} from those presented in \texttt{IFEval}~\citep{DBLP:journals/corr/abs-2311-07911}.
These constraints form the foundation of our synthetic data generation pipeline, loosely inspired by \texttt{Instruct-SkillMix}~\citep{DBLP:journals/corr/abs-2408-14774}, which (1) proposes new general-purpose prompts, (2) assigns valid combinations of verifiable constraints along with the parameters associated with each constraint and (3) generates new prompts that incorporate mixtures of the verifiable constraints in natural language over a wide variety of domains (e.g., blog post and cooking recipe).
Using this pipeline, we obtain a total of 48K unique synthetic prompts which contain mixtures of four, five or six verifiable constraints.

Using the synthetic prompts, we then apply two commonly-used methods for automatically curating preference pairs: rejection sampling (RS) and Monte Carlo Tree Search (MCTS).
Rejection sampling presents a straightforward method to extract preference pairs for a given prompt by generating $N$ independent responses with the policy model, scoring each response and using (high, low) scoring pairs of responses as the preference pairs~\citep{DBLP:conf/icml/YuanPCLSXW24, DBLP:conf/naacl/KhakiLMYR24}.
On the other hand, Monte Carlo Tree Search presents a more complex method for extracting preference pairs -- for a given tree obtained via MCTS, where each node represents the partial response generated for the given prompt, any pair of sibling nodes with (high, low) scoring pairs of responses are used as preference pairs~\citep{DBLP:journals/corr/abs-2405-00451, DBLP:journals/corr/abs-2406-03816}.
The RS approach is computationally efficient but there is no structure to the generated responses.
Meanwhile, the MCTS approach is more resource intensive, but returns pairs of responses that share common prefixes and a more nuanced contrast in their remaining suffixes.

We use our synthetic prompts and the associated preference pairs to systematically investigate how different heuristics used for automatically curating preference pairs impact models' downstream performances.
To this end, we focus on three critical dimensions that characterize a preference dataset:
\begin{enumerate}
    \item \textbf{Shared prefixes in preference pairs}: Does structural consistency (e.g., common prefixes) between chosen and rejected responses improve learning?
    \item \textbf{Contrast and quality of responses}: Is high-contrast or low-contrast pairing always superior, or does a mix of high- and low-contrast pairs offer better results?
    \item \textbf{Difficulty of training prompts}: How does the complexity of training prompts affect the model’s ability to generalize across different tasks?
\end{enumerate}


Our findings reveal several actionable insights:

\textbf{1. Preference pairs with shared prefixes (MCTS) marginally outperforms preference pairs that do not (RS) consistently over different training configurations.}
The performance of the MCTS-generated preference pairs is also more stable across different training configurations than the RS-generated preference pairs -- this stability is particularly valuable when response correctness is challenging to quantify.

\textbf{2. Having only high-contrast preference pairs is better than having only low-contrast preference pairs, but having a mixture of both often provides the best performance by balancing learning efficiency and diversity.}
Our results also consistently indicate that the relative \textit{contrast} between the chosen and rejected responses have a greater impact than their absolute correctness.

\textbf{3. Training on moderately difficult prompts results in better generalization across evaluation tasks, including more complex ones.}
Curating preference datasets with excessively challenging prompts returns a low yield rate of preference pairs due to the lower success rate, and even given the same dataset size, the difficulty of the prompts tend to overwhelm the model and hinder the learning efficiency.

\begin{figure}
    \centering
    \includegraphics[scale=0.71, clip, trim=0.6cm 6.2cm 1.5cm 1.0cm]{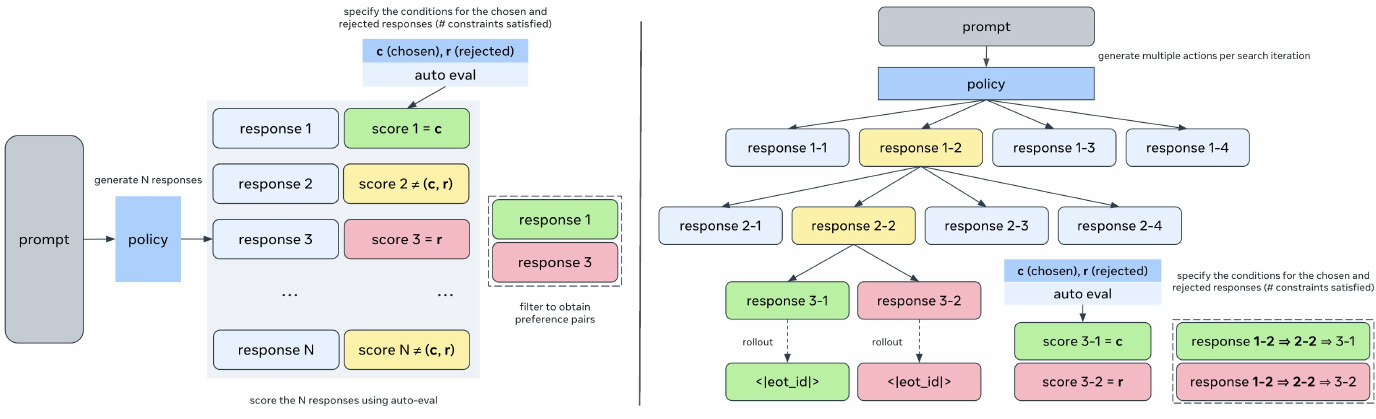}
    \caption{
    Automatically curating preference pairs via rejection sampling (RS, left) and Monte Carlo Tree Search (MCTS, right).
    \textbf{RS}: We independently sample $N$ different outputs from the policy, score each output with a verifier and take (high, low) scoring responses as the (chosen, rejected) pairs.
    \textbf{MCTS}: We perform tree search with the policy while generating multiple actions per each search iteration.
    Then, we use the rollouts from sibling nodes with (high, low) reward scores as the (chosen, rejected) pairs to obtain preference pairs with common prefixes up to the parent nodes.
    }
    \label{fig:rs-mcts}
\end{figure}

\section{Background}
\label{section:background}
\textbf{Preference Learning.}
Preference learning is a technique that is used to align LLMs during their post-training phase, involving pairs of (chosen, rejected) responses in the training dataset.
It aligns LLMs by steering them towards generating the chosen responses and away from generating the rejected responses.
The Bradley-Terry model~\citep{DBLP:journals/corr/bradleyterry1952} provides the probabilistic framework for preference learning by modeling the pairwise comparison between two responses ($y_1$, $y_2$) provided by the LLM to a given prompt $x$:
$$p(y_1 \succ y_2 | x) = \frac{\text{exp}(r^*(x, y_1))}{\text{exp}(r^*(x, y_1)) + \text{exp}(r^*(x, y_2))}$$
Common methods for preference learning such as Direct Preference Optimization (DPO,~\cite{DBLP:conf/nips/RafailovSMMEF23}) directly optimize the model to update its parameters to increase the likelihood of generating the chosen response over the rejected response.
Meanwhile, other methods such as Proximal Policy Optimization (PPO,~\cite{DBLP:journals/corr/SchulmanWDRK17}) indirectly perform this optimization by first training a reward model to assign scores corresponding to the preferences in the training data, and then optimizing a policy model with the guidance of the reward model.
Both approaches have been instrumental in aligning LLMs with human preferences and enhancing their capabilities for a wide array of downstream tasks.

\textbf{Data Curation for Preference Learning.} 
The success of preference learning for LLM post-training has naturally led researchers to propose methods for automatically curating preference pairs designed to further boost model capabilities~\citep{DBLP:conf/icml/YuanPCLSXW24, DBLP:conf/naacl/KhakiLMYR24, DBLP:journals/corr/abs-2405-00451}.
Such methods assign scores to LLM-generated outputs using verifiers to determine which outputs should be preferred during training. 
While various other methods have been proposed for curating preference data~\citep{DBLP:journals/corr/abs-2402-11411, DBLP:journals/corr/abs-2404-02078, DBLP:journals/corr/abs-2406-18629}, in this paper we focus on two popular methods: rejection sampling (RS) and Monte Carlo Tree Search (MCTS).
During rejection sampling, the policy model generates $N$ independent responses to the given prompt and a verifier scores each response according to some evaluation metric or a reward model.
Responses with (high, low) scores are selected as the (chosen, rejected) responses, respectively.
Meanwhile, in the MCTS framework, the policy model performs tree search for the given prompt by generating a fixed number of tokens at each iteration and builds the tree with nodes that represent each subsequence of generated tokens.
During MCTS, the policy model performs rollouts by generating full responses and backpropagates the reward scores assigned to the rollouts.
This results in a tree of possible responses generated for a given prompt, with each node being assigned a Q-value which measures the quality of the response generated so far.
Here, sibling nodes with sufficient differences in Q-values or reward scores are selected such that the preference pairs contain common prefixes and the suffixes account for the quality difference between the two responses.

While such methods return preference pairs that are effective for alignment and capability improvements, there is a lack of studies into exactly \textit{how} the preference pairs should be curated based on these methods.
In this work we perform a systematic investigation of how different characteristics of preference datasets affect downstream performance of LLMs, using instruction following accompanied with verifiable constraints as our task of interest.
We choose instruction-following as our task in order to incorporate multiple constraints into the prompt and score the response on a fine-grained level based on the ratio of constraints that are satisfied. 
We use verifiable constraints to assign quality scores to our responses in a reliable and efficient manner.

\section{Prompt Synthesis}
\label{section:prompt_synthesis}
\begin{table}
    \small
    \centering
    \begin{tabularx}{\textwidth}{llX}
        \toprule
        constraint & keyword args & description \\ \midrule
        \texttt{alliteration} & \texttt{num\_alliteration\_words} & the response should contain an alliteration, i.e. a sequence of X words starting with the same letter, where X $=$ \texttt{num\_alliteration\_words}.\\
        \texttt{ascending\_num\_words} & \texttt{n/a} & the response should contain sentences such that the number of words in each sentence is in ascending order.\\
        \texttt{max\_word\_length} & \texttt{max\_word\_length} & the maximum length of all words in the response should be at most X characters, where X $=$ \texttt{max\_word\_length}.\\
        \texttt{num\_words\_per\_sentence} & \texttt{relation}, \texttt{num\_words} & each sentence in the response should contain R $\in \{\text{at least, at most}\}$ X words, where R $=$ \texttt{relation} and X $=$ \texttt{num\_words}.\\
        \texttt{required\_sentence} & \texttt{sentence} & the response should contain a sentence S, where S $=$ \texttt{sentence}. \\
        \texttt{tldr\_summary} & \texttt{n/a} & the response should end with a “TL;DR” on a new line summarizing the response.\\
        \bottomrule
    \end{tabularx}
    \caption{
    Examples of verifiable constraints used for our synthetic prompts.
    Some constraints require one or more keyword arguments that materialize the constraint for its associated prompt, while others do not require any argument.
    }
    \label{tab:constraint_examples}
\end{table}

We perform all our experiments with a new set of synthetic prompts that incorporate mixtures of verifiable constraints.
These prompts allow us to evaluate the qualities of the generated responses in both a consistent and fine-grained manner, providing a suitable environment for us to control the attributes of the preference dataset and investigate their impact on downstream performance.

\subsection{Instruction-Following with Verifiable Constraints}
\label{subsection:constraint-ontology}
Our verifiable constraints resemble but are distinct from the set of constraints provided in \texttt{IFEval}~\citep{DBLP:journals/corr/abs-2311-07911}.
We define 23 constraints which can be deterministically verified using code, spanning ones that check for adherence to specific structural, stylistic, or formatting requirements.
The complete ontology of our constraints is provided in Table~\ref{tab:constraint_all} in the appendix.

We design our verifiable constraints such that they follow the formatting of the verifiable constraints provided in \texttt{IFEval} for unified evaluation -- each constraint is accompanied by a set of keyword arguments that are needed to actualize the constraint for the given prompt.
Refer to Table~\ref{tab:constraint_examples} for examples of such constraints and their associated keyword arguments.
For example, the \texttt{alliteration} constraint is paired with a keyword argument \texttt{num\_alliteration\_words}, which indicates the number of words that must display the alliteration.
Some constraints such as \texttt{tl;dr\_summary} do not contain any keyword arguments as they are self-explanatory and do not need any further specifications.

\begin{figure}
    \centering
    \includegraphics[scale=0.67, clip, trim=0.8cm 0.6cm 1.2cm 3.0cm]{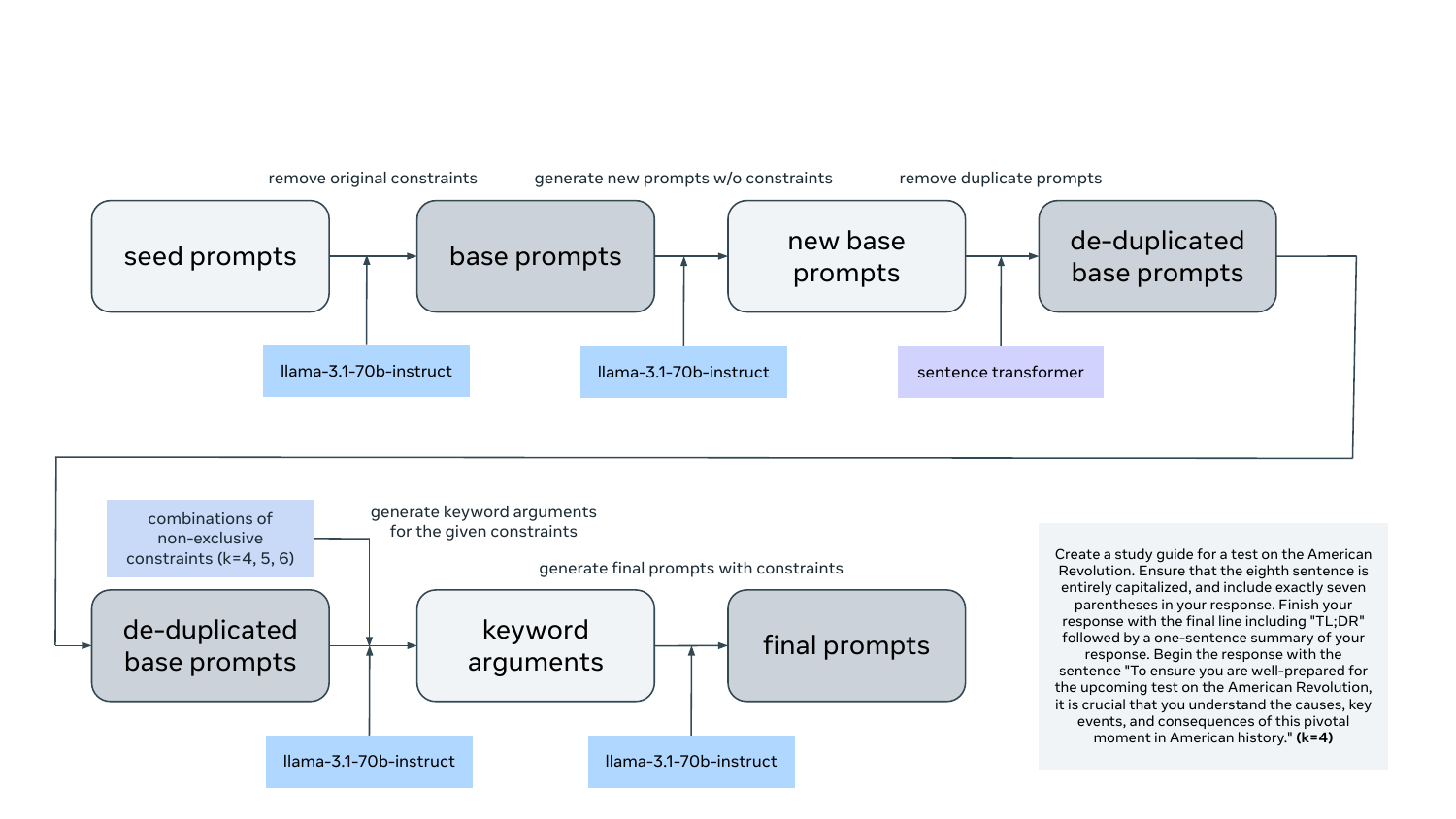}
    \caption{
    Overview of our pipeline for generating synthetic prompts with verifiable constraints.
    We first take a set of seed prompts from an existing dataset where the prompts contain constraints, and remove all constraints with an LLM (\texttt{llama-3.1-70b-instruct}) to obtain base prompts corresponding to the original dataset.
    Next, we randomly sample a small subset of the base prompts and use them as few-shot examples to generate new prompts \textit{without any constraints}.
    We remove duplicates among the newly-generated prompts and the existing base prompts using a sentence transformer.
    Then, we randomly sample a combination of $k \in \{4,5,6\}$ of our verifiable constraints that are non-conflicting and use an LLM to generate the input parameters required for the set of selected constraints.
    Finally, we use the resulting input kwargs and the new base prompts to generate the final prompts that integrate the constraints in natural language.
    }
    \label{fig:prompt-synthesis}
\end{figure}

\subsection{Prompt Generation}
\label{subsection:prompt-generation}
We build a pipeline that generates synthetic prompts for instruction-following, incorporating combinations of different verifiable constraints to create a set of challenging prompts that stress-test models' capabilities of following complex instructions.
Our pipeline is similar to \texttt{Instruct-SkillMix}~\citep{DBLP:journals/corr/abs-2408-14774} in terms of mixing skills, except that we have a preexisting set of verifiable constraints functioning as the ``skills'' and incorporate additional layers for generating the keyword arguments before generating the final instructions.
In terms of scalability, our pipeline does not require human supervision and generates synthetic prompts only using \texttt{llama-3.1-70b-instruct}~\citep{DBLP:journals/corr/abs-2407-21783}.

Figure~\ref{fig:prompt-synthesis} provides a visual summary of our synthetic prompt generation pipeline.
Our data generation pipeline can be decomposed into five main processes.
First, we take a set of seed instruction-following prompts (e.g., from \texttt{IFEval}) and remove the additional constraints originally associated with the dataset.
We few-shot prompt \texttt{llama-3.1-70b-instruct} with examples of additional constraints being removed and obtain a set of prompts from the seed dataset without constraints.
Second, we take the base prompts and use \texttt{llama-3.1-70b-instruct} to generate new prompts in their base forms, taking an approach similar to \texttt{Self-Instruct}~\citep{DBLP:conf/acl/WangKMLSKH23} by using the existing base prompts as few-shot examples and prompting the model to generate 20 new prompts at a time.
Third, we remove the newly generated prompts that are semantically duplicates either with any of the seed set of base prompts or any of the other new prompts.
We use \texttt{all-mpnet-base-v2}~\citep{DBLP:conf/nips/Song0QLL20}, a lightweight sentence transformer, to compute semantic embeddings and use the dot product to compute similarity scores.
Fourth, we randomly sample a combination of $k$ constraints which do not contradict each other, and for each constraint in this mixture, we randomly sample or generate the associated keyword arguments.
We perform random sampling for keyword arguments that can be randomly chosen without considering the prompt such as the \texttt{relation} argument in the \texttt{num\_words\_per\_sentence} constraint or the \texttt{num\_alliteration\_words} argument in the \texttt{alliteration} constraint.
Meanwhile, we use \texttt{llama-3.1-70b-instruct} to generate the keyword arguments that require more contextual understanding of the prompt, such as the \texttt{sentence} argument for the \texttt{required\_sentence} constraint.
Fifth, we take the set of constraints along with the keyword arguments chosen for each prompt and use \texttt{llama-3.1-70b-instruct} to rewrite the base prompts into their final forms which integrate the constraints and their keyword arguments in natural language.

We use our data generation pipeline to generate synthetic prompts with mixtures of verifiable constraints such that we can score the quality of any arbitrary response to the given prompt by using the constraints and their keyword arguments along with a deterministic evaluation code.
In contrast to \texttt{IFEval} which integrates at most $k=3$ constraints into the instruction, our pipeline allows us to integrate any number of constraints to create synthetic prompts, resulting in more challenging prompts and leaving room for a more diverse array of quality of responses in our preference datasets.
We use $k \in \{4,5,6\}$ in our experiments.
Our choice of using relatively higher values of $k$ is to make our synthetic training prompts maximally distinct from \texttt{IFEval}~\citep{DBLP:journals/corr/abs-2311-07911} for more reliable evaluation, and to perform evaluations on challenging, out-of-distribution test prompts with equally high values of $k$.
Refer to Table~\ref{tab:prompt_examples} in the appendix for examples of our synthetic prompts.

\subsection{Prompt Information}
\label{subsection:prompt-info}
We generate instruction-following prompts for combinations of $k\in\{4, 5, 6\}$ constraints.
Table~\ref{tab:prompt_statistics} shows the statistics that describe the prompts that we generate.
For each value of $k$, we generate about 16K synthetic prompts, resulting in a total of about 48K prompts being used in our experiments.
The average length of the prompts increases as the number of constraints increases, as the complexity of the prompts increases with larger numbers of constraints.

Table~\ref{tab:prompt_examples} displays examples of synthetic prompts generated using our pipeline for a given combination of verifiable constraints.
Each prompt contains a general-purpose base instruction such as writing a short story or a speech, and is accompanied by a combination of verifiable constraints defined in our ontology with keyword arguments that satisfy the context of the instruction -- for example, the first example uses the word "shouting" for the \texttt{nth\_sent\_first\_word} constraint, which is appropriate for the context of the base instruction involving a boy lost in a shopping mall.
For any arbitrary response to a given prompt, we can use the associated constraints and keyword arguments to automatically assign a score indicating whether the response follows each constraint and aggregate the scores to assign an overall correctness score to the response.

\begin{table}
    \centering
    \resizebox{\textwidth}{!}{
    \begin{tabular}{ccc|ccc|ccc}
    \toprule
        \multicolumn{3}{c}{$k=4$} & \multicolumn{3}{c}{$k=5$} & \multicolumn{3}{c}{$k=6$} \\\midrule
        {\scriptsize \texttt{num\_prompts}} & {\scriptsize \texttt{mean\_words}} & {\scriptsize \texttt{std\_words}} & {\scriptsize \texttt{num\_prompts}} & {\scriptsize \texttt{mean\_words}} & {\scriptsize \texttt{std\_words}} & {\scriptsize \texttt{num\_prompts}} & {\scriptsize \texttt{mean\_words}} & {\scriptsize \texttt{std\_words}} \\\midrule
        \small 15,900 & \small 70.62 & \small 14.37 & \small 15,739 & \small 84.17 & \small 15.61 & \small 15,559 & \small 97.95 & \small 16.91 \\
    \bottomrule
    \end{tabular}
    }
    \caption{
    Statistics of synthetic prompts generated by our pipeline.
    We generate about 16K prompts for each value of $k$, resulting in 48K prompts total across all our experiments.
    Note that \texttt{num\_prompts} refers to the number of prompts, \texttt{mean\_words} refers to the average number of words in each prompt, and \texttt{std\_words} refers to the standard deviation of the number of words in each prompt.
    The number of words in each prompt increases with the number of constraints.
    }
    \label{tab:prompt_statistics}
\end{table}

\begin{table}
    \centering
    \small
    \begin{tabularx}{\textwidth}{p{5.1cm}X}
        \toprule
        constraints & prompt \\\midrule
        \texttt{ascending\_num\_words}, \texttt{freq\_long\_words}, \texttt{max\_word\_length}, \texttt{nth\_sent\_first\_word}, \texttt{start\_checker} & Write a short story about a boy who gets lost in a shopping mall. Include at least 7 words that are at least 12 characters long, and ensure that the sentences have an increasing number of words, i.e. each sentence should contain more words than its previous one. Also, only include words that are at most 12 characters long. Make sure that the fifth sentence starts with the word "shouting", and begin your response with the sentence "As the sounds of loud chatter and clinking of dishes filled the food court, little Tommy suddenly discovered that his parents were nowhere to be seen.". \\
        \\
        \texttt{nth\_sent\_first\_word}, \texttt{num\_bold\_words}, \texttt{num\_exclamations}, \texttt{tldr\_summary}, \ \texttt{vowel\_capitalization} & Write a motivational speech for a high school graduation ceremony. Capitalize the vowels in your response, and include seven words that are bolded in HTML format (e.g., <b>word</b>). Also, ensure that the sixth sentence starts with the word "today". Make sure that the response contains exactly three exclamation marks, and finish the response with the final line including "TL;DR" followed by a one-sentence summary of your response. \\
        \bottomrule
    \end{tabularx}
    \caption{
    Examples of synthetic prompts generated for $k=5$, which contains a combination of five verifiable constraints.
    }
    \label{tab:prompt_examples}
\end{table}

\section{Preference Data Curation}
\label{section:preference_curation}
Using the prompts obtained in Section~\ref{section:prompt_synthesis}, we generate responses and extract preference data using the aggregated correctness scores of the responses.
Here, we assign high-scoring responses as the chosen responses and low-scoring responses as the rejected responses.
We employ two methods for preference data curation: rejection sampling (RS) and Monte Carlo Tree Search (MCTS).
Rejection sampling presents a straightforward and efficient way to obtain preference data but the (chosen, rejected) responses do not share any relationship.
On the other hand, MCTS is more expensive and slower to run, but returns (chosen, rejected) responses that share a prefix with more nuanced contrast.
We use two contrasting approaches to generate diverse types of preference data and additionally investigate the effect of having common prefixes in preference pairs.

\textbf{Rejection Sampling.}
Refer to Figure~\ref{fig:rs-mcts} (left) for a visual overview. 
We first set a filtering criteria, where the chosen response must achieve a score of $\mathbf{c}$ according to our verifier and the rejected response must achieve a score of $\mathbf{r}$.
Then, we generate $N$ different responses independently with the policy model and score each response with our verifier.
Given a prompt $x$, its associated set of verifiable constraints $\mathcal{C}$, the response $r$ and our verifier $\mathcal{V}$ which verifies whether the response satisfies any given constraint $c$, we compute the score as
$$R(r|x, \mathcal{C}) = \frac{1}{\|\mathcal{C}\|}\sum_{c\in\mathcal{C}}\mathcal{V}(r|x, c)$$
We extract all preference pairs such that 1) the chosen score equals $\mathbf{c}$, 2) the rejected score equals $\mathbf{r}$, and 3) there are no overlapping responses between the extracted preference pairs.
Using rejection sampling offers a simple method for automatically curating preference pairs, but it returns pairs that are independently sampled with no common structures between the chosen and rejected responses.

\textbf{Monte Carlo Tree Search.}
Refer to Figure~\ref{fig:rs-mcts} (right) for a visual overview.
We conduct Monte Carlo Tree Search (MCTS) with a granularity level of token sequences -- for a given prompt $x$ and the MCTS tree $T$, each node $s_i$ in $T$ represents a partial response generated for $x$, and each edge $(s_i, s_j)$ in $T$, also known as an \textit{action}, represents a sequence of tokens generated from $s_i \rightarrow s_j$.
In this setup, we use an LLM $\Pi$ as the policy and our verifier as the outcome reward model via three major steps: selection, expansion and backpropagation.

\paragraph{Selection.}
Given the current node $s_t$ and $K$ different actions $(a_1, ..., a_K)$ generated by the policy from $s_t$, we balance exploitation and exploration to select the next node for tree search.
Each action is a sequence of tokens under a pre-specified maximum length.

Our selection depends on $Q(s_t, a)$ and $N(s_t, a)$, the Q-value and visit count of each subsequent node reached by taking action $a$ from $s_t$, respectively.
We use the Predictor+Upper Confidence bounds applied to Trees (PUCT) and select the next node $s_{t+1}$ according to the following formula:
$$s_{t+1}^* = \argmax_{(s_{t+1} = s_t\rightarrow a_i)}\left[Q(s_t,a_i)+c_{\text{puct}}\cdot \Pi(a_i|s_t)\frac{\sqrt{N(s_t)}}{1 + N(s_t, a_i)}\right]$$
Refer to Appendix~\ref{sec:mcts_details} for more information on how to compute the policy score $\Pi(a_i|s_t)$.
Using PUCT, we prioritize exploration during the initial stages of tree building when the visit counts have low values, and prioritize exploitation during the later stages of tree building as the visit counts increase and effectively weigh the Q-values more for scoring.
This results in a balanced trade-off during our tree search.

\paragraph{Expansion.}
We perform expansion from the current node $s_t$ and generate $K$ new actions with the policy $\Pi$.
For each new action, we perform $M$ separate rollouts and score each rollout using a linear combination of the score $\mathcal{V}$ assigns and the self-evaluation score assigned by $\Pi$.
We use self-evaluation, denoted as $\Pi_{\text{self-eval}}(s_t)$, in addition to the verifier scores, to provide step-level feedback during the tree search. Again, refer to Appendix~\ref{sec:mcts_details} for more information on how to compute $\Pi_{\text{self-eval}}(s_t)$.
$$R(s_t) = (1 - \lambda)\cdot \left[\frac{1}{M\|\mathcal{C}\|}\sum_{i\in [1...M]}\sum_{c\in\mathcal{C}}\mathcal{V}(\Pi_{\text{rollout}}(s_t)|x, c)\right] + \lambda \cdot \Pi_{\text{self-eval}}(s_t)$$
After assigning the reward scores to each rollout, we average the scores across the rollouts for each new action $a_i$ for $i \in [1,\ldots,K]$ and add the node $s_{t+1} = (s_t\rightarrow a_i)$, with the averaged reward score, to the tree.

\paragraph{Backpropagation.}
After computing the reward scores for the rollouts and adding new nodes, we backpropagate the reward scores through the parent nodes.
We increment the visit counts and directly update the Q-values of a given (state, action) pair based on the Q-values and the visit counts of the children nodes of $s_{t+1} = (s_t, a)$, as the following:
$$N(s_t) = N(s_t) + 1$$
$$Q(s_{t}, a) = \frac{\sum_{i=1}^KQ(s_{t+1}, a_i)\cdot N(s_{t+1}, a_i) + R(s_{t+1})}{\sum_{i=1}^KN(s_{t+1}, a_i) + 1}$$

We repeat the threefold process over multiple iterations for each root node, and we traverse down the tree while switching the root node until we reach a terminal node.

Once the trees are constructed, we curate preference data from pairs of sibling nodes, i.e. nodes that share the same parent node.
For pairs of non-leaf sibling nodes that satisfy the correctness criteria in the tree, we sample their rollouts to obtain complete responses.
We use the same criteria for our data curation as that of rejection sampling -- we set a filtering criteria in which the chosen response must score $\mathbf{c}$ and the rejected response must score $\mathbf{r}$, and sample all pairs with no overlapping responses.
Note that we use the scores assigned by the verifier $\mathcal{V}$ only, and not the policy's self-evaluation, in order to ensure that the correctness of the (chosen, rejected) responses stay consistent across the RS- and MCTS-based curation methods.

\begin{figure}
    \centering
    \includegraphics[scale=0.72, clip, trim=0.1cm 8.2cm 4.5cm 0.1cm]{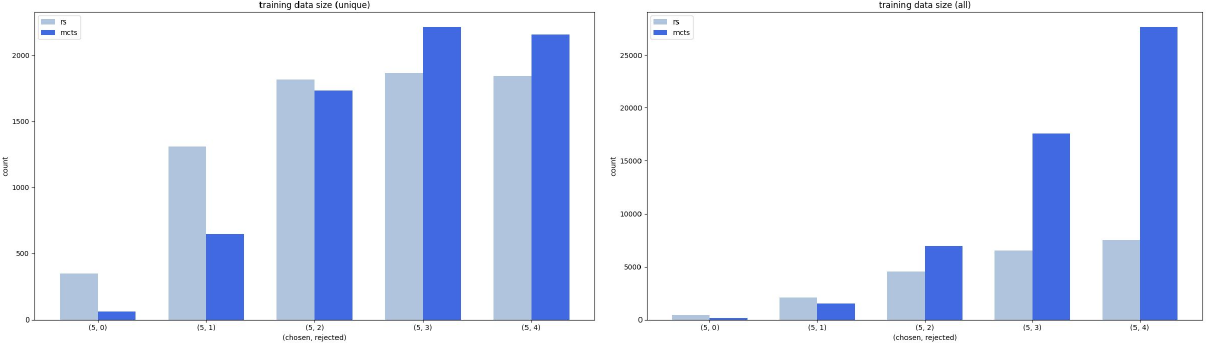}
    \caption{
    Number of preference pairs for different correctness filtering criteria at $k=5$.
    The light blue color indicates preference pairs obtained via rejection sampling (RS), and the dark blue color indicates preference pairs obtained via Monte Carlo Tree Search (MCTS).
    The left subfigure shows the number of unique prompts with (chosen, rejected) responses associated with each filtering criteria, and the right subfigure shows the total number of preference pairs with (chosen, rejected) responses associated with each filtering criteria.
    }
    \label{fig:preference-pair-stats}
\end{figure}

\textbf{Preference Pair Statistics.}
We curate preference pairs using both RS- and MCTS-based methods for our synthetic prompts with $k \in \{4,5,6\}$.
Figure~\ref{fig:preference-pair-stats} shows the number of preference pairs obtained via both methods for $k=5$ when the correctness criteria for (chosen, rejected) pairs is set as (5 correct, 0 correct), (5 correct, 1 correct), (5 correct, 2 correct), (5 correct, 3 correct) and (5 correct, 4 correct) pairs.
The left subfigure depicts the number of unique prompts corresponding to the preference pairs extracted for each filtering criteria.
Using rejection sampling (RS) returns a higher yield of preference pairs with high contrast between the (chosen, rejected) responses, while using Monte Carlo Tree Search (MCTS) returns a higher yield of preference pairs with low contrast between the responses.
The right subfigure depicts the total number of preference pairs extracted for each criteria -- the same observation can be made about the relative yield, with MCTS yielding a large number of preference pairs due to its tree structure.
We use preference pairs collected for $k \in \{4,5,6\}$ using both methods over different filtering criteria to perform our experiments.

\section{Experiments and Results}
\label{section:experiments}
We perform experiments to systematically investigate the effects of preference dataset attributes on the downstream performance of language models.
To this end, we examine three critical dimensions of the preference dataset: the (1) presence of shared prefixes between the (chosen, rejected) responses, (2) contrast and quality of the chosen and rejected responses, and (3) difficulty or complexity of the training prompts.

\textbf{Data curation setup.}
We implement rejection sampling by generating $N=64$ outputs for each prompt in our training set with a temperature of 1.0, and score each output using the verifier $\mathcal{V}$.
Meanwhile, we implement Monte Carlo Tree Search with maximum depth of 5, number of actions 4, number of rollouts 4, $c_{puct} = 1.0$ and $\lambda = 0.2$.
We score all rollouts using the same verifier $\mathcal{V}$.

\textbf{Training setup.} 
We run our experiments with \texttt{llama-3.1-8b-instruct}~\citep{DBLP:journals/corr/abs-2407-21783} and finetune the model on the preference dataset using DPO~\citep{DBLP:conf/nips/RafailovSMMEF23}.
We finetune the model for one epoch and use a maximum sequence length of 2048, learning rate of \texttt{5e-7} with a linear scheduler and 4 gradient accumulation steps with a total batch size of 32.

\textbf{Evaluation setup.}
We evaluate our models on \texttt{IFEval}~\citep{DBLP:journals/corr/abs-2311-07911}, as well as three different synthetic evaluation sets that are designed to be more challenging than \texttt{IFEval}.
Our synthetic evaluation sets are created using the same pipeline described in Section~\ref{subsection:prompt-generation}, but with the verifiable constraints provided in \texttt{IFEval} to maintain the distinction between the constraints used for training and for evaluation.
We use $k\in \{4,5,6\}$ for our evaluation datasets and synthesize about 500 evaluation prompts for each value of $k$.

Using the finetuned models, we generate 16 responses to each prompt at temperature 0.7 and measure the average score.
We use two metrics to evaluate each response: 1) a hard score measuring whether the response satisfies \textit{all} the constraints, and 2) a soft score measuring the \textit{ratio} of the constraints satisfied by the response.

\setlength{\tabcolsep}{2.4pt}
\begin{table}
    \centering
    \scriptsize
    \begin{tabular}{lcccclcccc}
    \toprule
        \multicolumn{5}{c}{\textbf{training k=4}} & \multicolumn{5}{c}{\textbf{training k=5}}\\
        Method & IFEval & $\text{Ours}_{k=4}$ & $\text{Ours}_{k=5}$ & $\text{Ours}_{k=6}$ & Method & IFEval & $\text{Ours}_{k=4}$ & $\text{Ours}_{k=5}$ & $\text{Ours}_{k=6}$\\\midrule
        RS, (c=4, r=1) & 79.24 & 37.49 & 20.54 & 13.78 & RS, (c=5, r=2) & 76.32 & 35.70 & 19.46 & 12.61 \\
        MCTS, (c=4, r=1) & 79.48 & 39.16 & 20.39 & 14.11 & MCTS, (c=5, r=2) & 76.59 & 37.81 & 20.06 & 12.93 \\\midrule
        RS, (c=4, r=2) & 78.86 & 38.56 & 21.66 & 14.71 & RS, (c=5, r=3) & 76.25 & 35.28 & 18.91 & 12.24 \\
        MCTS, (c=4, r=2) & 79.68 & 39.22 & 22.43 & 15.75 & MCTS, (c=5, r=3) & 76.47 & 36.73 & 19.94 & 13.62 \\\midrule
        RS, (c=4, r=3) & 76.74 & 35.02 & 19.40 & 12.85 & RS, (c=5, r=4) & 74.18 & 33.38 & 17.35 & 10.38 \\
        MCTS, (c=4, r=3) & 79.59 & 39.05 & 21.61 & 14.96 & MCTS, (c=5, r=4) & 75.15 & 34.88 & 18.75 & 11.74 \\\midrule
        RS, (c=3, r=0) & 80.06 & 39.25 & 21.54 & 14.37 & RS, (c=4, r=1) & 78.95 & 39.40 & 21.74 & 14.58 \\
        MCTS, (c=3, r=0) & 79.39 & 39.15 & 21.56 & 13.61 & MCTS, (c=4, r=1) & 78.63 & 39.21 & 21.19 & 14.63 \\\midrule
        RS, (c=3, r=1) & 80.06 & 39.15 & 21.90 & 15.20 & RS, (c=4, r=2) & 78.07 & 37.02 & 20.17 & 13.05 \\
        MCTS, (c=3, r=1) & 79.94 & 39.23 & 22.15 & 14.93 & MCTS, (c=4, r=2) & 78.48 & 38.53 & 22.12 & 15.22 \\\midrule
        RS, (c=3, r=2) & 77.52 & 36.20 & 19.52 & 12.48 & RS, (c=4, r=3) & 75.64 & 32.98 & 17.83 & 11.19 \\
        MCTS, (c=3, r=2) & 77.89 & 38.89 & 21.51 & 13.97 & MCTS, (c=4, r=3) & 77.40 & 37.63 & 20.84 & 14.06 \\\midrule
        RS, (c=4, r=1/2/3) & 78.70 & 36.68 & 20.20 & 13.57 & RS, (c=4, r=1/2/3) & 79.08 & 37.98 & 20.50 & 13.42 \\
        MCTS, (c=4, r=1/2/3) & 79.97 & 39.31 & 22.19 & 15.89 & MCTS, (c=4, r=1/2/3) & 78.47 & 39.03 & 22.63 & 15.25 \\\midrule
        RS, (c=3/4, r=0/1/2/3) & 79.10 & 37.62 & 20.98 & 13.99 & RS, (c=4/5, r=0/1/2/3) & 78.18 & 36.81 & 19.72 & 13.29 \\
        MCTS, (c=3/4, r=0/1/2/3) & 79.42 & 39.37 & 22.24 & 15.20 & MCTS, (c=4/5, r=0/1/2/3) & 77.89 & 38.17 & 22.00 & 14.38 \\
    \bottomrule
    \end{tabular}
    \caption{
    Evaluation results comparing preference data without shared prefixes (RS) and with shared prefixes (MCTS).
    We show results for different training data configurations for $k\in\{4,5\}$.
    Each $(c=n_1,r=n_2)$ indicates that the chosen response correctly addresses $n_1$ constraints and the rejected response correctly addresses $n_2$ constraints.
    The results for $k=6$ and the soft score metrics for all experiments are provided in Tables~\ref{tab:common_prefix_full_results_k4},~\ref{tab:common_prefix_full_results_k5} and~\ref{tab:common_prefix_full_results_k6} in the appendix.
    }
    \label{tab:common_prefix_results}
\end{table}

\subsection{Shared Prefixes in Preference Pairs}
\label{subsection:common-prefix}
We investigate the effect of having structural consistency between the (chosen, rejected) responses in the preference dataset.
Recent techniques utilize tree search to curate fine-grained preference pairs where the (chosen, rejected) responses differ after a shared prefix -- we examine the effects of utilizing such preference pairs.
To this end, we use rejection sampling (RS), which returns responses without shared prefixes, and Monte Carlo Tree Search (MCTS), which returns responses with shared prefixes, to curate preference datasets under identical conditions.

\setlength{\tabcolsep}{5.5pt}
\begin{table}
    \centering
    \scriptsize
    \begin{tabular}{lcccclcccc}
    \toprule
        \multicolumn{5}{c}{\textbf{training k=4}} & \multicolumn{5}{c}{\textbf{training k=5}}\\
        Method & IFEval & $\text{Ours}_{k=4}$ & $\text{Ours}_{k=5}$ & $\text{Ours}_{k=6}$ & Method & IFEval & $\text{Ours}_{k=4}$ & $\text{Ours}_{k=5}$ & $\text{Ours}_{k=6}$\\\midrule
        RS, (c=3, r=0) & 79.04 & 39.45 & 20.86 & 13.82 & RS, (c=4, r=1) & 76.56 & 36.08 & 19.44 & 11.88 \\
        RS, (c=3, r=1) & 78.66 & 38.47 & 20.56 & 14.02 & RS, (c=4, r=2) & 74.57 & 34.09 & 17.95 & 10.93 \\
        RS, (c=3, r=2) & 74.60 & 33.59 & 18.46 & 11.23 & RS, (c=4, r=3) & 72.80 & 31.50 & 16.68 & 10.01 \\\midrule
        RS, (c=4, r=1) & 79.24 & 37.49 & 20.54 & 13.78 & RS, (c=5, r=2) & 76.32 & 35.70 & 19.46 & 12.61 \\
        RS, (c=4, r=2) & 77.44 & 37.69 & 20.33 & 13.77 & RS, (c=5, r=3) & 74.13 & 33.30 & 17.31 & 11.21 \\
        RS, (c=4, r=3) & 74.09 & 33.38 & 17.65 & 10.73 & RS, (c=5, r=4) & 72.40 & 30.72 & 16.26 & 9.51 \\\midrule
        MCTS, (c=3, r=0) & 78.30 & 37.86 & 20.52 & 12.81 & MCTS, (c=4, r=1) & 76.71 & 37.24 & 19.49 & 12.37 \\
        MCTS, (c=3, r=1) & 77.58 & 38.03 & 19.76 & 13.71 & MCTS, (c=4, r=2) & 75.23 & 35.65 & 18.74 & 12.17 \\
        MCTS, (c=3, r=2) & 75.05 & 34.96 & 18.69 & 11.53 & MCTS, (c=4, r=3) & 74.23 & 32.39 & 17.22 & 9.73 \\\midrule
        MCTS, (c=4, r=1) & 79.48 & 39.16 & 20.39 & 14.11 & MCTS, (c=5, r=2) & 76.59 & 37.81 & 20.06 & 12.93 \\
        MCTS, (c=4, r=2) & 77.76 & 38.25 & 20.00 & 14.26 & MCTS, (c=5, r=3) & 76.14 & 36.04 & 19.39 & 12.37 \\
        MCTS, (c=4, r=3) & 75.65 & 35.03 & 18.80 & 12.36 & MCTS, (c=5, r=4) & 74.06 & 32.14 & 17.81 & 10.52 \\
    \bottomrule
    \end{tabular}
    \caption{
    Evaluation results studying the effects of (chosen, rejected) response quality.
    We provide results for different training prompt difficulties across $k\in\{4,5\}$, as well as for both RS- and MCTS-based data curation methods.
    Each $(c=n_1,r=n_2)$ denotes that the chosen response correctly addresses $n_1$ constraints and the rejected response correctly addresses $n_2$ constraints.
    The soft score metrics for all experiments are provided in Tables~\ref{tab:response_quality_results_k4} and~\ref{tab:response_quality_results_k5} in the appendix.
    }
    \label{tab:response_quality_results}
\end{table}

We perform our experiments by fixing the size of the training dataset as well as the number of unique prompts associated with the preference pairs that occur in each training set, across our comparison experiments.
Meanwhile, we perform experiments by varying the other two dimensions: the correctness of the (chosen, rejected) responses, and the difficulty of the training prompts as measured by the values of $k$.

Table~\ref{tab:common_prefix_results} shows the results of our experiments.
We observe two key findings from the results.

\textbf{MCTS outperforms RS by a small margin consistently over different training configurations.}
While not a significant difference, we observe that models trained on preference datasets curated via MCTS slightly outperform models trained on preference datasets curated via RS across most of our training configurations.
However, this difference may not be significant enough to warrant the added complexity introduced by MCTS in regular settings unless the additional performance gains are necessary.

\textbf{MCTS offers more consistent performance over different training configurations than RS.}
We observe that models trained on preference datasets curated via MCTS demonstrate more consistent performance across the training configurations than ones trained on preference datasets curated via RS.
This implies that for cases where the constraint is not programmatically verifiable and the correctness of the response is more difficult to quantify, it may be more effective to use the preference dataset curated by MCTS as it offers more stable performance.

\subsection{Contrast and Quality of Responses}
\label{subsection:response-quality}
We investigate the effects of controlling the quality, or correctness, of the (chosen, rejected) responses in the preference dataset.
Across our experiments, we maintain the same training dataset size and the number of unique prompts given a fixed data curation method and training prompt difficulty.
Table~\ref{tab:response_quality_results} shows the results of varying the correctness of the (chosen, rejected) pairs across diverse training configurations including the RS/MCTS-based curation method, as well as the training prompt difficulties with $k=4$ or $5$.

Meanwhile, we also study the effects of having a mixture of both high- and low-contrast pairs.
To compare its performance to those of using only high- or low-contrast pairs, we fix the training dataset size and the number of unique prompts, and replace some of the rejected responses with low correctness scores with rejected responses that have slightly higher correctness scores.
For this experiment, we use MCTS for data curation while using $k=4$ or $5$ for the complexity of the training prompts.

\begin{figure}
    \centering
    \includegraphics[scale=0.6, clip, trim=0.1cm 1.6cm 0.2cm 0.1cm]{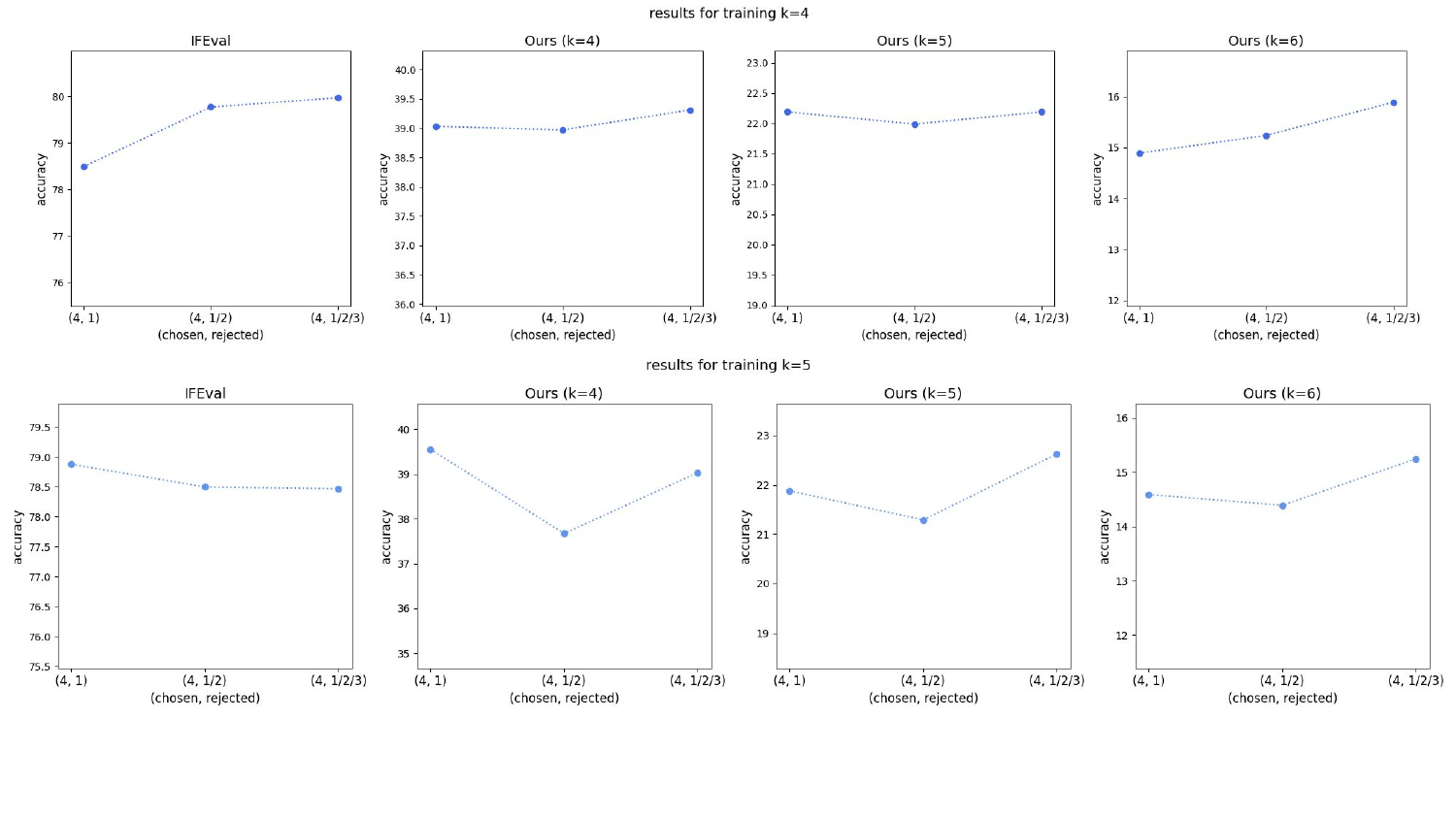}
    \caption{
    Evaluation results demonstrating the effects of mixing preference pairs with different margins between the (chosen, rejected) responses.
    The two rows correspond to our training setup with different values of $k$ (number of verifiable constraints in each prompt), and the four columns correspond to our evaluation sets.
    The x-axis indicates the correctness of the (chosen, rejected) responses with lower-margin pairs mixed in while keeping the same training size.
    The y-axis represents the accuracies.
    Results for more experiments are provided in Tables~\ref{tab:response_quality_results_mix_k4} and~\ref{tab:response_quality_results_mix_k5} in the appendix.
    }
    \label{fig:quality_mixture_results}
\end{figure}

The results of our experiments are visualized in Figure~\ref{fig:quality_mixture_results}.
We show the composition of the (chosen, rejected) responses in terms of their correctness on the x-axis, and the evaluation metric on the y-axis for each subplot.
For example, for a training set with $k=4$ constraints, we begin with a (chosen, rejected) correctness of (c, r) $=$ (4, 1) and then mix in r$=$2 to obtain (c, r) $=$ (4, 1/2), and then mix in r$=$3 to obtain (c, r) $=$ (4, 1/2/3).

We make three key observations from our experiments.

\textbf{When used alone, high-contrast preference pairs is more helpful than low-contrast preference pairs.}
Our results in Table~\ref{tab:response_quality_results} show that for a fixed correctness of the chosen response, increasing the correctness of the rejected response decreases the performance in a consistent manner.
For example, using a correctness of c = 3 for the chosen response and changing the correctness of the rejected response from r = (0, 1, 2) for a training dataset with $k=4$ decreases the \texttt{IFEval} score from 79.04$\rightarrow$78.66$\rightarrow$74.60.
This trend holds across both RS- and MCTS-based data curation methods, as well as across the difficulty of the training prompts.
Likewise, we find that using high-contrast preference pairs is better than using low-contrast pairs for downstream performance.

\textbf{The margin between the (chosen, rejected) responses have a bigger impact than the absolute scores themselves.}
Table~\ref{tab:response_quality_results} demonstrates that preference datasets with the same value of $c-r$ for the correctness of the (chosen, rejected) responses result in similar performances across our evaluation sets.
For example, on the \texttt{IFEval} benchmark our results for rejection sampling with (c, r) $=(3, 0)$ and (c, r) $=(4, 1)$ with a training set of $k=4$ result in scores of 79.04 vs. 79.24, while our results for rejection sampling with (c, r) = $(4, 1)$ and (c, r) = $(5, 2)$ with a training set of $k=5$ result in scores of 76.56 vs. 76.32.
Our results indicate that the margin between the preference pairs have more influence on the downstream performance than the absolute correctness of the (chosen, rejected) responses as long as the chosen responses are reasonably correct.

\textbf{Having a mixture of both high-contrast and low-contrast pairs sometimes demonstrates better performance than only using high-contrast pairs.}
Our experiments shown in Figure~\ref{fig:quality_mixture_results} indicate that given the same training dataset size, mixing high-contrast and low-contrast preference pairs sometimes return better performance than only using high contrast pairs.
For example, in a case where there are $k=4$ constraints in the training set, a model trained on preference pairs with (c, r) = (4, 1) scores 78.49 on \texttt{IFEval}, but models trained on preference pairs with (c, r) = (4, 1/2) and (4, 1/2/3) score 79.77 and 79.97, respectively.
However, in another case where there are $k=5$ constraints in the training set, a model trained on preference pairs with (c, r) = (4, 1) scores 78.88 on \texttt{IFEval}, but models trained on preference pairs with (c, r) = (4, 1/2) and (4, 1/2/3) score 78.50 and 78.47, not showing improvements.
Our additional experiments in Tables~\ref{tab:response_quality_results_mix_k4} and~\ref{tab:response_quality_results_mix_k5} in the appendix also indicate that having a mixture of contrasts often helps, but the results are too mixed to yield a definitive conclusion.

\subsection{Training Prompt Difficulty}
\label{subsection:prompt-difficulty}
We examine how the difficulty of prompts provided in a preference dataset affects downstream performance across evaluation sets of varying difficulties.
Similar to previous experiments, we control the size of the training dataset and the number of unique prompts to compare across the prompt difficulties.
For each experiment, we fix the preference data curation method and the margin between the (chosen, rejected) responses while comparing between the three complexities of the prompts in our training set with $k \in \{4, 5, 6\}$.
Then, we repeat our experiments across our data curation methods and the qualities of the (chosen, rejected) responses.

\setlength{\tabcolsep}{5.5pt}
\begin{table}
    \centering
    \scriptsize
    \begin{tabular}{lcccclcccc}
    \toprule
        \multicolumn{5}{c}{\textbf{Rejection Sampling (RS)}} & \multicolumn{5}{c}{\textbf{Monte Carlo Tree Search (MCTS)}}\\
        Method & IFEval & $\text{Ours}_{k=4}$ & $\text{Ours}_{k=5}$ & $\text{Ours}_{k=6}$ & Method & IFEval & $\text{Ours}_{k=4}$ & $\text{Ours}_{k=5}$ & $\text{Ours}_{k=6}$\\\midrule
        $k=4$, (c=3, r=0) & 79.70 & 38.41 & 21.15 & 14.27 & $k=4$, (c=3, r=0) & 78.94 & 39.00 & 21.60 & 14.40 \\
        $k=5$, (c=4, r=1) & 79.51 & 38.66 & 21.48 & 14.70 & $k=5$, (c=4, r=1) & 78.72 & 39.42 & 21.43 & 14.25 \\
        $k=6$, (c=5, r=2) & 77.83 & 36.42 & 20.15 & 12.28 & $k=6$, (c=5, r=2) & 78.13 & 38.55 & 21.64 & 14.45 \\\midrule
        $k=4$, (c=3, r=1) & 79.36 & 39.22 & 22.00 & 14.78 & $k=4$, (c=3, r=1) & 79.12 & 38.81 & 21.72 & 15.66 \\
        $k=5$, (c=4, r=2) & 78.58 & 37.70 & 21.42 & 13.61 & $k=5$, (c=4, r=2) & 77.53 & 38.93 & 21.69 & 15.20 \\
        $k=6$, (c=5, r=3) & 76.73 & 36.16 & 19.96 & 12.87 & $k=6$, (c=5, r=3) & 77.18 & 37.06 & 20.16 & 12.84 \\
    \bottomrule
    \end{tabular}
    \caption{
    Evaluation results investigating the effects of training prompt difficulty ($k\in \{4, 5, 6\}$) on downstream performance for evaluation sets of varying difficulties.
    We provide results for both RS- and MCTS-based preference data curation methods, as well as for different margins between the (chosen, rejected) responses.
    Each $(c=n_1,r=n_2)$ indicates that the chosen response correctly addresses $n_1$ constraints and the rejected response correctly addresses $n_2$ constraints.
    The soft score metrics for all experiments are provided in Tables~\ref{tab:prompt_difficulty_full_results_rs} and~\ref{tab:prompt_difficulty_full_results_mcts} in the appendix.
    }
    \label{tab:prompt_difficulty_results}
\end{table}

Table~\ref{tab:prompt_difficulty_results} summarizes the results of our experiments.
We observe two key findings from the results.

\textbf{Training on moderately difficult prompts is overall more helpful than training on extremely difficult prompts.}
We find that models trained on moderately difficult prompts perform better than models trained on extremely difficult prompts in all our evaluation sets.
For example, our model trained on preference pairs for prompts with $k=4$ scores 79.70 and 79.36 on \texttt{IFEval} with (c, r) = (3, 0) and (3, 1), respectively, while the same model trained on preference pairs for prompts with $k=6$ scores 77.83 and 76.73 on \texttt{IFEval} with (c, r) = (5, 2) and (5, 3), respectively.
The same pattern holds across other evaluation sets and across both data curation methods, showcasing the importance of moderating prompt complexities for preference learning.

\textbf{Training on moderately difficult prompts is more helpful even for performing well on extremely difficult prompts at test time.}
Models trained on moderately complex prompts ($k=4$) outperform models trained on extremely complex prompts ($k=6$) even for evaluation sets involving $k=6$ constraints.
For example, our model trained on prompts of $k=4$ with preference pairs obtained via RS with (c, r) = (3, 0) scores 14.27 on our evaluation set with $k=6$ while our models trained on prompts of $k=6$ with preference pairs obtained via RS with (c, r) = (5, 2) scores 12.28 on the same evaluation set.
Our results indicate that it is better to use training prompts of moderate difficulties to achieve generalization to more difficult prompts.

\subsection{Additional Experiments}
We perform two additional experiments to confirm that preference learning helps our models gain instruction-following skills, and investigate the limitations of our methods.
To this end, we (1) compare the performances of SFT and DPO to ensure that our preference datasets teach meaningful skills to our models, and (2) examine how our RS-based preference data curation method scales with varying amounts of compute.

\setlength{\tabcolsep}{2.0pt}
\begin{table}
    \centering
    \scriptsize
    \begin{tabular}{lcccclcccc}
    \toprule
        \multicolumn{5}{c}{\textbf{Rejection Sampling (RS)}} & \multicolumn{5}{c}{\textbf{Monte Carlo Tree Search (MCTS)}}\\
        Method & IFEval & $\text{Ours}_{k=4}$ & $\text{Ours}_{k=5}$ & $\text{Ours}_{k=6}$ & Method & IFEval & $\text{Ours}_{k=4}$ & $\text{Ours}_{k=5}$ & $\text{Ours}_{k=6}$\\\midrule
        policy (\texttt{llama-3.1-8b-instruct}) & 71.71 & 27.77 & 14.15 & 7.34 & - & - & - & - & - \\\midrule
        SFT, $k=4$, (c=4, r=1/2/3) & 74.77 & 28.17 & 15.10 & 7.10 & SFT, $k=4$, (c=4, r=1/2/3) & 73.30 & 29.70 & 15.02 & 7.50 \\
        DPO, $k=4$, (c=4, r=1/2/3) & 78.70 & 36.68 & 20.20 & 13.57 & DPO, $k=4$, (c=4, r=1/2/3) & 79.97 & 39.31 & 22.19 & 15.89 \\\midrule
        SFT, $k=5$, (c=4, r=1/2/3) & 74.75 & 28.36 & 14.53 & 7.74 & SFT, $k=5$, (c=4, r=1/2/3) & 73.87 & 28.54 & 13.49 & 6.68 \\
        DPO, $k=5$, (c=4, r=1/2/3) & 79.08 & 37.98 & 20.50 & 13.42 & DPO, $k=5$, (c=4, r=1/2/3) & 78.47 & 39.03 & 22.63 & 15.25 \\
    \bottomrule
    \end{tabular}
    \caption{
    Evaluation results comparing the performance of training on our preference datasets via DPO compared to the base policy model or running SFT on the chosen responses only.
    We provide results for both RS- and MCTS-based preference data curation methods, as well as for different training prompt difficulties ($k\in \{4,5\}$).
    Each $(c=n_1,r=n_2)$ indicates that the chosen response correctly addresses $n_1$ constraints and the rejected response correctly addresses $n_2$ constraints.
    The soft score metrics for all experiments are provided in Tables~\ref{tab:sft_dpo_full_results_rs} and~\ref{tab:sft_dpo_full_results_mcts} in the appendix.
    }
    \label{tab:sft_dpo_results}
\end{table}

\textbf{Does preference learning actually help?}
We demonstrate that our preference learning datasets teach meaningful instruction-following skills to the models and allows us to perform the comparison experiments shown in Sections~\ref{subsection:common-prefix},~\ref{subsection:response-quality} and~\ref{subsection:prompt-difficulty}.
To this end, we compare the performances of the models trained via DPO with the base policy (\texttt{llama-3.1-8b-instruct}) and the policy trained via supervised fine-tuning (SFT) on the chosen responses only.
For SFT, we train each model for three epochs with a maximum sequence length of 2048, learning rate of \texttt{2e-6} with a linear scheduler and 4 gradient accumulation steps with a total batch size of 32.

Table~\ref{tab:sft_dpo_results} shows the results of our experiments.
First, we confirm that DPO significantly improves over the baseline policy model even though the verifiable constraints in the training set and the evaluation sets are separate -- this indicates that training on a set of verifiable constraints can be helpful for generalizing to another set of verifiable constraints as long as they involve transferable skills.
Moreover, we observe that DPO outperforms SFT across all training configurations, demonstrating the importance of our preference learning setup with the rejected responses in addition to the chosen responses.

\textbf{How does rejection sampling scale?}
We investigate how different values of $N$ during rejection sampling affect its overall quality and ensure that our RS-based method is designed to be competitive against MCTS.
To this end, we re-run the rejection sampling curation pipeline for $N\in\{4, 8, 16, 32, 64\}$ for a fixed difficulty $k=5$ and correctness scores of (c, r) = (4, 1) and (4, 2) and measure the performance across all of our evaluation sets.

\begin{figure}
    \centering
    \includegraphics[scale=0.6, clip, trim=0.1cm 2.2cm 0.2cm 0.1cm]{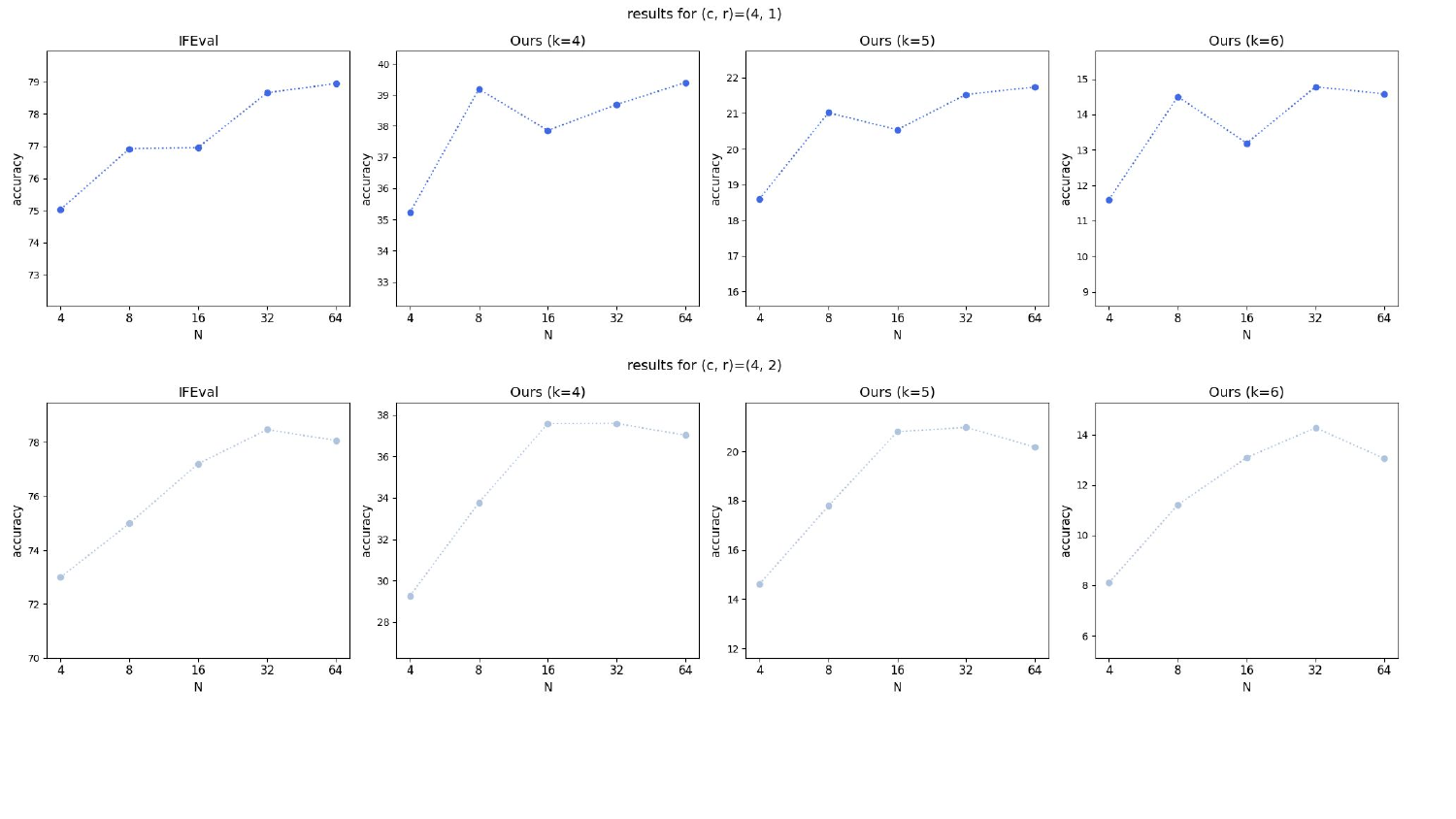}
    \caption{
    Evaluation results for increasing the number of outputs generated for rejection sampling (RS) from $N=4$ to $N=64$, given a set of training prompts with $k=5$ and (c, r) = (4, 1) or (4, 2).
    We observe a steady improvement in performance as more outputs are generated per prompt until $N=32$, where it begins to saturate or even deteriorate.
    }
    \label{fig:baseline-ablations}
\end{figure}

Figure~\ref{fig:baseline-ablations} shows the results of our experiments.
We observe that the performance of the models trained on preference pairs curated with RS-based methods increases with $N$, but also saturate around $N=32$ and do not observe significant performance improvements afterward.
This indicates that curating preference data via rejection sampling, while efficient and straightforward to scale, has limitations that cannot be solved by simply scaling the number of generated outputs, and further improvements may require more sophisticated search strategies such as the MCTS-based data curation method implemented in this work.

\section{Conclusion}
\label{section:conclusion}
We systematically investigate the effects of various attributes of preference datasets on model capabilities from the perspective of instruction-following.
To this end, we first build a data generation pipeline that combines general-purpose prompts with mixtures of verifiable constraints to synthesize challenging instruction-following prompts.
We then automatically curate preference pairs using two popular methods: rejection sampling (RS) and Monte Carlo Tree Search (MCTS).
Using the preference pairs, we examine the effects of (1) the existence of shared prefixes between the chosen and rejected responses, (2) the contrast and quality of the responses, and (3) the complexity of the training prompts.
Our results indicate that having a common prefix in the preference pairs offers marginal yet consistent improvements, high-contrast preference pairs outperform low-contrast pairs but a mixture is sometimes better than both, and training on moderately difficult prompts is more helpful than training on extremely difficult prompts.
Our work provides a systematic framework for curating different types of preference datasets and sets the groundwork for future studies that extend the scope beyond verifiable instruction-following constraints to more general constraints.



\clearpage
\newpage
\bibliographystyle{assets/plainnat}
\bibliography{paper}

\clearpage
\newpage
\beginappendix
\label{section:appendix}
\section{Complete Ontology of Verifiable Constraints and Training Examples}
We present our 23 verifiable constraints in Table~\ref{tab:constraint_all}, and examples of our synthetic prompts in Table~\ref{tab:prompt_examples}.
\begin{table}[!h]
    \small
    \centering
    \begin{tabularx}{\textwidth}{lp{3.6cm}X}
        \toprule
        constraint & keyword args & description \\ \midrule
        \texttt{alliteration} & \texttt{num\_alliteration\_words} & the response should contain an alliteration, i.e. a sequence of X words starting with the same letter, where X $=$ \texttt{num\_alliteration\_words}.\\
        \texttt{ascending\_num\_words} & \texttt{n/a} & the response should contain sentences such that the number of words in each sentence is in ascending order.\\
        \texttt{edit\_response} & \texttt{n/a} & the response should be separated by dashes ‘------’ separating two responses with the second response improving upon the first response.\\
        \texttt{end\_quotation} & \texttt{n/a} & the last sentence should be wrapped in quotation marks.\\
        \texttt{first\_letter\_capital} & \texttt{n/a} & the first letter of each word should be capitalized.\\
        \texttt{frequency\_long\_words} & \texttt{relation}, \texttt{num\_words}, \texttt{word\_length} & the response should contain R $\in \{\text{at least, at most}\}$ X words that are at least Y characters long each, where R $=$ \texttt{relation}, X $=$ \texttt{num\_words} and Y $=$ \texttt{word\_length}.\\
        \texttt{keywords\_ordered} & \texttt{keywords} & the response should contain a set of keywords in the given order and not in any other order.\\
        \texttt{max\_word\_length} & \texttt{max\_word\_length} & the maximum length of all words in the response should be at most X characters, where X $=$ \texttt{max\_word\_length}.\\
        \texttt{no\_period} & \texttt{n/a} & the response should contain no periods.\\
        \texttt{nth\_sentence\_capital} & \texttt{nth\_sentence} & the \texttt{nth\_sentence} in the response should be all capitalized (and only the nth sentence).\\
        \texttt{nth\_sentence\_first\_word} & \texttt{first\_word}, \texttt{num\_sentences}, \texttt{nth\_sentence} & the first word of the \texttt{nth\_sentence} should be the given word W in the instruction, where W $=$ \texttt{first\_word}.\\
        \texttt{num\_words\_per\_sentence} & \texttt{relation}, \texttt{num\_words} & each sentence in the response should contain R $\in \{\text{at least, at most}\}$ X words, where R $=$ \texttt{relation} and X $=$ \texttt{num\_words}.\\
        \texttt{number\_bold\_words} & \texttt{num\_words} & the response should bold exactly X words in HTML format <b>word</b>, where X $=$ \texttt{num\_words}.\\
        \texttt{number\_exclamations} & \texttt{relation}, \texttt{num\_exclamations} & the response should contain R $\in \{\text{at least, at most}\}$ X exclamation marks, where R $=$ \texttt{relation} and X $=$ \texttt{num\_exclamations}.\\
        \texttt{number\_italic\_words} & \texttt{num\_words} & the response should italicize exactly X words in textile format ‘\_word\_’, where X $=$ \texttt{num\_words}.\\
        \texttt{number\_parentheses} & \texttt{num\_parentheses} & the response should contain exactly X parentheses (), where X $=$ \texttt{num\_parentheses}.\\
        \texttt{number\_parts} & \texttt{part\_splitter}, \texttt{num\_parts} & the response should contain X parts such as “Part 1”, “PART 1”, “Part 2”, “PART 2”, ... where \texttt{part\_splitter} $\in \{\text{Part, PART}\}$ and X $=$ \texttt{num\_parts}.\\
        \texttt{numbered\_headers} & \texttt{num\_headers} & the response should contain X enumerated headings starting with 1. , 2. , 3. , ... , where X $=$ \texttt{num\_headers}..\\
        \texttt{required\_sentence} & \texttt{sentence} & the response should contain a sentence S, where S $=$ \texttt{sentence}.\\
        \texttt{start\_checker} & \texttt{first\_sentence} & the response should begin with \texttt{first\_sentence}.\\
        \texttt{tldr\_summary} & \texttt{n/a} & the response should end with a “TL;DR” on a new line summarizing the response.\\
        \texttt{variable\_placeholder\_format} & \texttt{relation}, \texttt{num\_placeholders} & the response should contain R $\in \{\text{at least, at most}\}$ X variable placeholders in curly brackets, where R $=$ \texttt{relation} and X $=$ \texttt{num\_placeholders}.\\
        \texttt{vowel\_capitalization} & \texttt{n/a} & the response should capitalize all vowels in the response.\\
        \bottomrule
    \end{tabularx}
    \caption{
    Complete list of verifiable constraints used for our synthetic prompts.
    }
    \label{tab:constraint_all}
\end{table}

\clearpage
\begin{table*}[!h]
    \centering
    \small
    \begin{tabularx}{\textwidth}{p{3.5cm}X}
        \toprule
        constraints & prompt \\\midrule
        \texttt{ascending\_num\_words}, \texttt{freq\_long\_words}, \texttt{max\_word\_length}, \texttt{nth\_sent\_first\_word}, \texttt{start\_checker} & Write a short story about a boy who gets lost in a shopping mall. Include at least 7 words that are at least 12 characters long, and ensure that the sentences have an increasing number of words, i.e. each sentence should contain more words than its previous one. Also, only include words that are at most 12 characters long. Make sure that the fifth sentence starts with the word "shouting", and begin your response with the sentence "As the sounds of loud chatter and clinking of dishes filled the food court, little Tommy suddenly discovered that his parents were nowhere to be seen.". \\
        \\
        \texttt{nth\_sent\_first\_word}, \texttt{num\_bold\_words}, \texttt{num\_exclamations}, \texttt{tldr\_summary}, \ \texttt{vowel\_capitalization} & Write a motivational speech for a high school graduation ceremony. Capitalize the vowels in your response, and include seven words that are bolded in HTML format (e.g., <b>word</b>). Also, ensure that the sixth sentence starts with the word "today". Make sure that the response contains exactly three exclamation marks, and finish the response with the final line including "TL;DR" followed by a one-sentence summary of your response. \\
        \\
        \texttt{alliteration}, \texttt{keywords\_ordered}, \texttt{nth\_sent\_first\_word}, \texttt{number\_bold\_words}, \texttt{vowel\_capitalization} & Write a scene where a character walks into a room and is surprised by what they see. Capitalize the vowels in your response and include exactly six words that are bolded in HTML format (e.g., <b>word</b>). Include an alliteration of five consecutive words beginning with the same letter of the alphabet, and include the words 'door', 'space', 'chaos' in your response in the exact order provided. Moreover, make sure that the fourth sentence starts with the word "suddenly".\\
        \\
        \texttt{alliteration}, \texttt{number\_italic\_words}, \texttt{number\_parts}, \texttt{required\_sentence}, \texttt{start\_checker} & Write a short essay on the importance of vitamin D in human health. Start your response with the sentence "Vitamin D is an essential nutrient that has been gaining increasing attention in recent years, primarily due to its significant role in preventing various health issues, ranging from bone diseases to certain cancers." Include the sentence "Research has shown that vitamin D deficiency is associated with an increased risk of cardiovascular disease, diabetes, and osteoporosis." in your response, and include one word that is italicized in textile format, wrapped between underscore characters (e.g., \_word\_). Divide your essay into one part marked with 'Part 1', and include an alliteration of five consecutive words beginning with the same letter of the alphabet.\\
        \\
        \texttt{edit\_response}, \texttt{num\_exclamations}, \texttt{start\_checker}, \texttt{tldr\_summary}, \texttt{vowel\_capitalization} & Write a short script of a news anchor introducing a breaking news story. Capitalize the vowels in your response, provide two responses separated by six plus signs ++++++, and include at least nine exclamation marks. Also, include a "TL;DR" summarizing the breaking news story at the end of your response. Begin the response with the sentence "We interrupt your regular programming to bring you this breaking news story coming in from the White House.".\\
        \\
        \texttt{alliteration}, \texttt{ascending\_num\_words}, \texttt{nth\_sentence\_capital}, \texttt{number\_italic\_words}, \texttt{number\_parentheses} & Create a data model for a medical records database, including at least one relationship and one constraint. Write the first sentence of the response in all capital letters, include eight words that are italicized in textile format, and include an alliteration of five consecutive words beginning with the same letter of the alphabet. Ensure that the sentences in the response have an increasing number of words, i.e. each sentence should contain more words than its previous one. Include exactly six parentheses in the response.\\
        \\
        \texttt{max\_word\_length}, \texttt{number\_parentheses}, \texttt{number\_parts}, \texttt{tldr\_summary}, \texttt{vowel\_capitalization} & Write a list of 10 ways to improve your public speaking skills. Capitalize the vowels in the response. Have one part marked with PART 1. Finish your response with the final line including "TL;DR" followed by a one-sentence summary of your response. Only include words that are at most 14 characters long, and include exactly eight parentheses in the response.\\
        \\
        \texttt{alliteration}, \texttt{nth\_sentence\_first\_word}, \texttt{num\_words\_per\_sentence}, \texttt{number\_bold\_words}, \texttt{vowel\_capitalization} & Write a letter to a historical figure, asking for their advice on a modern issue. Ensure that the vowels in the response are capitalized and there are eight words that are bolded in HTML format (e.g., <b>word</b>). Include an alliteration of three consecutive words beginning with the same letter of the alphabet. Make sure that the sixth sentence starts with the word "nonetheless", and that each sentence in the response is less than 15 words long.\\
        \bottomrule
    \end{tabularx}
    \caption{
    Examples of synthetic prompts generated for $k=5$, which combines five verifiable constraints.
    }
    \label{tab:prompt_examples}
\end{table*}

\clearpage
\section{MCTS Details}
\label{sec:mcts_details}
\textbf{Computing the policy score.}
We compute the policy score $\Pi(a_i|s_t)$ by computing the average of the log probabilities of tokens generated for action $a_i$ from state $s_t$, with the denominator moderated by a hyperparameter $\gamma$, which we set to 1.0 for our experiments.
Refer to the formula below for the exact definition:
$$\Pi(a_i|s_t) = \text{exp}\left[\frac{1}{\|a_i\|^{\gamma}}\sum_{j=1}^{\|a_i\|}\Pi(t_j|s_t, t_{1...j-1})\right]$$
Note that $t_j$ denotes the $j$th token in action $a_i$.

\textbf{Computing the self-evaluation score.}
We compute the self-evaluation score $\Pi_{\text{self-eval}}$ by prompting the policy with a self-evaluation prompt $P_{\text{self-eval}}$, coupled with the response generated by the model so far, and obtaining the log probabilities of the final token of the response, denoted as $t_{\text{final}} \in \{\text{yes}, \text{no}\}$.
We use self-consistency~\citep{Wang2022SelfConsistencyIC} to obtain multiple self-evaluations of the policy of its own output over $L$ generations and average the scores in order to improve the reliability of our self-evaluation scores via increased compute.
$$\Pi_{\text{self-eval}} = \frac{1}{L}\sum_{i=1}^{L}\frac{1 + \text{exp}(\Pi(t_{\text{final}}=\text{yes}|P_{\text{self-eval}}, s_t)) - \text{exp}(\Pi(t_{\text{final}}=\text{no}|P_{\text{self-eval}}, s_t))}{2}$$
Our formula allows us to normalize the score between 0 to 1, with 0.5 indicating a neutral state where the confidence scores for $\Pi(t_{\text{final}}=\text{yes}|P_{\text{self-eval}}, s_t) = \Pi(t_{\text{final}}=\text{no}|P_{\text{self-eval}}, s_t)$.
We provide the self-evaluation prompt $P_{\text{self-eval}}$ in Figure~\ref{fig:self-evaluation-prompt}.

\section{Full Experiment Results}
\subsection{Common Prefix Results}
We provide the full results of our experiments investigating the effects of common prefixes in preference pairs in Tables~\ref{tab:common_prefix_full_results_k4},~\ref{tab:common_prefix_full_results_k5} and~\ref{tab:common_prefix_full_results_k6}.

\setlength{\tabcolsep}{8pt}
\begin{table}[!h]
    \centering
    \scriptsize
    \begin{tabular}{lcccc}
    \toprule
        & \multicolumn{4}{c}{\textbf{training k=4}} \\
        Method & IFEval & $\text{Ours}_{k=4}$ & $\text{Ours}_{k=5}$ & $\text{Ours}_{k=6}$ \\\midrule
        RS, (c=4, r=1) & 79.24 / 85.78 & 37.49 / 77.50 & 20.54 / 72.06 & 13.78 / 70.23 \\
        MCTS, (c=4, r=1) & 79.48 / 85.92 & 39.16 / 78.11 & 20.39 / 72.04 & 14.11 / 70.10 \\\midrule
        RS, (c=4, r=2) & 78.86 / 85.74 & 38.56 / 78.02 & 21.66 / 72.34 & 14.71 / 70.00 \\
        MCTS, (c=4, r=2) & 79.68 / 86.36 & 39.22 / 78.20 & 22.43 / 72.64 & 15.75 / 70.61 \\\midrule
        RS, (c=4, r=3) & 76.74 / 83.92 & 35.02 / 76.25 & 19.40 / 71.00 & 12.85 / 68.63 \\
        MCTS, (c=4, r=3) & 79.59 / 86.30 & 39.05 / 77.86 & 21.61 / 72.28 & 14.96 / 70.22 \\\midrule
        RS, (c=3, r=0) & 80.06 / 86.69 & 39.25 / 78.55 & 21.54 / 72.44 & 14.37 / 70.32 \\
        MCTS, (c=3, r=0) & 79.39 / 85.86 & 39.15 / 78.02 & 21.56 / 72.26 & 13.61 / 69.42 \\\midrule
        RS, (c=3, r=1) & 80.06 / 86.42 & 39.15 / 78.35 & 21.90 / 72.01 & 15.20 / 70.09 \\
        MCTS, (c=3, r=1) & 79.94 / 86.37 & 39.23 / 78.37 & 22.15 / 72.34 & 14.93 / 70.00 \\\midrule
        RS, (c=3, r=2) & 77.52 / 84.27 & 36.20 / 76.86 & 19.52 / 70.96 & 12.48 / 68.68 \\
        MCTS, (c=3, r=2) & 77.89 / 84.85 & 38.89 / 78.02 & 21.51 / 71.93 & 13.97 / 69.58 \\\midrule
        RS, (c=4, r=1/2/3) & 78.70 / 85.48 & 36.68 / 76.78 & 20.20 / 70.71 & 13.57 / 68.45 \\
        MCTS, (c=4, r=1/2/3) & 79.97 / 86.51 & 39.31 / 78.22 & 22.19 / 72.14 & 15.89 / 70.36 \\\midrule
        RS, (c=3/4, r=0/1/2/3) & 79.10 / 85.72 & 37.62 / 77.59 & 20.98 / 71.28 & 13.99 / 68.65 \\
        MCTS, (c=3/4, r=0/1/2/3) & 79.42 / 85.84 & 39.37 / 78.38 & 22.24 / 72.48 & 15.20 / 70.26 \\
    \bottomrule
    \end{tabular}
    \caption{
    Evaluation results comparing preference data without common prefixes (RS) and with common prefixes (MCTS).
    We show results for different training data configurations for $k=4$.
    Each $(c=n_1,r=n_2)$ indicates that the chosen response correctly addresses $n_1$ constraints and the rejected response correctly addresses $n_2$ constraints.
    The left score indicates the hard score (i.e., the proportion of responses that get \textit{all} constraints correct), and the right score indicates the soft score (i.e., the proportion of all constraints that are satisfied by the responses).
    }
    \label{tab:common_prefix_full_results_k4}
\end{table}

\newpage
\begin{figure}[!h]

\centering
\tcbset{colback=verylightgray, colframe=verylightgray, boxrule=0.5pt, arc=4pt}
\begin{tcolorbox}
\lstset{
    basicstyle=\ttfamily\color{black}\scriptsize,
    keywordstyle=\color{black},
    identifierstyle=\color{black},
    commentstyle=\color{black},
    stringstyle=\color{black},
    breaklines=true,
    backgroundcolor=\color{verylightgray},
    frame=none,
    numbers=none
}
\begin{lstlisting}
<|start_header_id|>user<|end_header_id|>

Evaluate whether the assistant's (partial) response to the given instruction follows the conditions specified in the instruction so far and does not violate any of the conditions. Complete the evaluation by using the words "yes" or "no", followed by an explanation for why the assistant's response follows or does not follow the given instruction so far.

Do NOT evaluate the conditions that can be checked automatically with Python code, including the ones listed below.
- DON'T EVALUATE: number of paragraphs/sentences/words/sections
- DON'T EVALUATE: existence of certain phrases/words/characters
- DON'T EVALUATE: capital or lowercase
Make sure to only evaluate the conditions that can be checked so far. For example, you cannot check if the response contains at least 20 sentences- this is because the given response is a partial, incomplete response and the full response later may possibly contain at least 20 sentences. Also, this can be checked automatically, so it corresponds to the first "DO NOT" condition listed above.

Instead, focus on the conditions that cannot be checked automatically and is more related to the content itself, as listed below.
- DO EVALUATE: whether the response follows the topic so far
- DO EVALUATE: whether the response matches the description of the characters/location/theme/etc laid out in the instruction so far
- DO EVALUATE: whether the response follows the tone requested in the instruction (e.g., persuasive, solemn, lively, etc.)
For example, if the response asks to write a conversation between a software engineer and a research scientist, make sure that there are two characters who are each software engineer and research scientist, respectively.

Instruction: %s
Response so far: %s

Begin your response by listing such content-based conditions and analyzing whether each condition has been satisfied on separate lines.
Be generous in terms of the evaluation criteria - only say "no" when you are sure that the partial response does not adhere to the content-based conditions. Otherwise, answer "yes" to each condition.
Most importantly, make sure to finish your evaluation with the phrase "Based on these evaluations, my overall evaluation is: ", followed by either "yes" or "no".

<|eot_id|><|start_header_id|>assistant<|end_header_id|>
\end{lstlisting}
\end{tcolorbox}
\caption{Our self-evaluation prompt $P_{\text{self-eval}}$.
We allow the policy to focus on the soft content of the response rather than the hard constraints that are verifiable by code.}
\label{fig:self-evaluation-prompt}
\end{figure}

\begin{table}[H]
    \centering
    \scriptsize
    \begin{tabular}{lcccc}
    \toprule
        & \multicolumn{4}{c}{\textbf{training k=5}} \\
        Method & IFEval & $\text{Ours}_{k=4}$ & $\text{Ours}_{k=5}$ & $\text{Ours}_{k=6}$ \\\midrule
        RS, (c=5, r=2) & 76.32 / 83.55 & 35.70 / 76.64 & 19.46 / 71.51 & 12.61 / 69.39 \\
        MCTS, (c=5, r=2) & 76.59 / 83.81 & 37.81 / 77.58 & 20.06 / 72.01 & 12.93 / 69.85 \\\midrule
        RS, (c=5, r=3) & 76.25 / 83.63 & 35.28 / 76.23 & 18.91 / 70.03 & 12.24 / 68.75 \\
        MCTS, (c=5, r=3) & 76.47 / 83.84 & 36.73 / 77.10 & 19.94 / 71.82 & 13.62 / 69.52 \\\midrule
        RS, (c=5, r=4) & 74.18 / 81.79 & 33.38 / 75.10 & 17.35 / 70.05 & 10.38 / 67.54 \\
        MCTS, (c=5, r=4) & 75.15 / 82.73 & 34.88 / 76.06 & 18.75 / 71.06 & 11.74 / 69.03 \\\midrule
        RS, (c=4, r=1) & 78.95 / 85.57 & 39.40 / 78.60 & 21.74 / 72.48 & 14.58 / 70.02 \\
        MCTS, (c=4, r=1) & 78.63 / 85.49 & 39.21 / 78.33 & 21.19 / 72.24 & 14.63 / 70.07 \\\midrule
        RS, (c=4, r=2) & 78.07 / 84.87 & 37.02 / 77.35 & 20.17 / 71.02 & 13.05 / 68.48 \\
        MCTS, (c=4, r=2) & 78.48 / 85.24 & 38.53 / 77.91 & 22.12 / 72.16 & 15.22 / 69.69 \\\midrule
        RS, (c=4, r=3) & 75.64 / 82.97 & 32.98 / 75.32 & 17.83 / 69.95 & 11.19 / 67.40 \\
        MCTS, (c=4, r=3) & 77.40 / 84.28 & 37.63 / 77.41 & 20.84 / 71.72 & 14.06 / 69.42 \\\midrule
        RS, (c=4, r=1/2/3) & 79.08 / 85.56 & 37.98 / 77.83 & 20.50 / 71.48 & 13.42 / 68.86 \\
        MCTS, (c=4, r=1/2/3) & 78.47 / 85.25 & 39.03 / 78.06 & 22.63 / 72.49 & 15.25 / 70.05 \\\midrule
        RS, (c=4/5, r=0/1/2/3) & 78.18 / 85.05 & 36.81 / 77.06 & 19.72 / 71.13 & 13.29 / 68.67 \\
        MCTS, (c=4/5, r=0/1/2/3) & 77.89 / 84.82 & 38.17 / 77.81 & 22.00 / 72.19 & 14.38 / 69.63 \\
    \bottomrule
    \end{tabular}
    \caption{
    Evaluation results comparing preference data without common prefixes (RS) and with common prefixes (MCTS).
    We show results for different training data configurations for $k=5$.
    Each $(c=n_1,r=n_2)$ indicates that the chosen response correctly addresses $n_1$ constraints and the rejected response correctly addresses $n_2$ constraints.
    The left score indicates the hard score (i.e., the proportion of responses that get \textit{all} constraints correct), and the right score indicates the soft score (i.e., the proportion of all constraints that are satisfied by the responses).
    }
    \label{tab:common_prefix_full_results_k5}
\end{table}

\begin{table}[H]
    \centering
    \scriptsize
    \begin{tabular}{lcccc}
    \toprule
        & \multicolumn{4}{c}{\textbf{training k=5}} \\
        Method & IFEval & $\text{Ours}_{k=4}$ & $\text{Ours}_{k=5}$ & $\text{Ours}_{k=6}$ \\\midrule
        RS, (c=6, r=2) & 75.17 / 82.74 & 33.45 / 75.19 & 17.00 / 69.84 & 10.19 / 67.39 \\
        MCTS, (c=6, r=2) & 75.70 / 83.11 & 34.06 / 75.71 & 17.07 / 70.27 & 10.46 / 67.64 \\\midrule
        RS, (c=6, r=3) & 77.47 / 84.44 & 36.99 / 77.22 & 20.05 / 71.16 & 13.27 / 68.39 \\
        MCTS, (c=6, r=3) & 78.06 / 85.04 & 38.68 / 77.96 & 21.63 / 72.21 & 14.94 / 70.02 \\\midrule
        RS, (c=6, r=4) & 76.09 / 83.51 & 35.68 / 75.63 & 19.07 / 71.01 & 11.99 / 68.47 \\
        MCTS, (c=6, r=4) & 76.26 / 83.58 & 35.81 / 75.63 & 18.41 / 71.05 & 11.56 / 68.64 \\\midrule
        RS, (c=5, r=1) & 75.13 / 82.77 & 34.17 / 75.56 & 18.48 / 70.93 & 11.93 / 68.49 \\
        MCTS, (c=5, r=1) & 76.09 / 83.47 & 36.34 / 76.52 & 19.04 / 71.33 & 12.70 / 68.62 \\\midrule
        RS, (c=5, r=2) & 75.01 / 82.68 & 34.48 / 75.88 & 18.10 / 70.64 & 11.21 / 68.60 \\
        MCTS, (c=5, r=2) & 76.57 / 83.83 & 36.03 / 76.72 & 18.75 / 70.93 & 11.62 / 68.68 \\\midrule
        RS, (c=5, r=3) & 76.73 / 83.89 & 36.16 / 76.62 & 19.96 / 70.97 & 12.87 / 68.15 \\
        MCTS, (c=5, r=3) & 77.18 / 84.53 & 37.06 / 77.10 & 20.16 / 71.25 & 12.84 / 68.57 \\\midrule
        RS, (c=5, r=1/2/3) & 78.10 / 84.97 & 37.66 / 77.61 & 21.45 / 71.49 & 13.69 / 68.84 \\
        MCTS, (c=5, r=1/2/3) & 78.33 / 85.17 & 40.12 / 78.34 & 23.12 / 72.32 & 15.28 / 69.73 \\\midrule
        RS, (c=5/6, r=1/2/3/4) & 78.48 / 85.28 & 37.82 / 77.86 & 22.73 / 72.02 & 14.39 / 69.48 \\
        MCTS, (c=5/6, r=1/2/3/4) & 79.01 / 85.72 & 40.58 / 78.93 & 23.04 / 73.18 & 16.00 / 70.57 \\
    \bottomrule
    \end{tabular}
    \caption{
    Evaluation results comparing preference data without common prefixes (RS) and with common prefixes (MCTS).
    We show results for different training data configurations for $k=6$.
    Each $(c=n_1,r=n_2)$ indicates that the chosen response correctly addresses $n_1$ constraints and the rejected response correctly addresses $n_2$ constraints.
    The left score indicates the hard score (i.e., the proportion of responses that get \textit{all} constraints correct), and the right score indicates the soft score (i.e., the proportion of all constraints that are satisfied by the responses).
    }
    \label{tab:common_prefix_full_results_k6}
\end{table}

\subsection{Response Quality Results}
\textbf{(chosen, rejected) response quality (unmixed).} We provide the full results of our experiments investigating the effects of the response correctness (or quality) in preference pairs in Tables~\ref{tab:response_quality_results_k4} and~\ref{tab:response_quality_results_k5}.

\textbf{(chosen, rejected) response quality (mixed).} We provide the full results of our experiments examining the effects of mixing preference pairs with different margins between the (chosen, rejected) responses in Tables~\ref{tab:response_quality_results_mix_k4} and~\ref{tab:response_quality_results_mix_k5}.

\setlength{\tabcolsep}{11pt}
\begin{table}[!h]
    \centering
    \scriptsize
    \begin{tabular}{lcccc}
    \toprule
        & \multicolumn{4}{c}{\textbf{training k=4}}\\
        Method & IFEval & $\text{Ours}_{k=4}$ & $\text{Ours}_{k=5}$ & $\text{Ours}_{k=6}$\\\midrule
        RS, (c=3, r=0) & 79.04 / 85.87 & 39.45 / 78.30 & 20.86 / 72.23 & 13.82 / 70.45\\
        RS, (c=3, r=1) & 78.66 / 85.51 & 38.47 / 78.15 & 20.56 / 72.37 & 14.02 / 70.13\\
        RS, (c=3, r=2) & 74.60 / 82.39 & 33.59 / 75.58 & 18.46 / 70.62 & 11.23 / 68.15\\\midrule
        RS, (c=4, r=1) & 79.24 / 85.78 & 37.49 / 77.50 & 20.54 / 72.06 & 13.78 / 70.23\\
        RS, (c=4, r=2) & 77.44 / 84.43 & 37.69 / 77.68 & 20.33 / 72.06 & 13.77 / 70.04\\
        RS, (c=4, r=3) & 74.09 / 82.16 & 33.38 / 75.30 & 17.65 / 70.21 & 10.73 / 67.79\\\midrule
        MCTS, (c=3, r=0) & 78.30 / 85.19 & 37.86 / 77.42 & 20.52 / 71.92 & 12.81 / 69.30\\
        MCTS, (c=3, r=1) & 77.58 / 84.69 & 38.03 / 77.74 & 19.76 / 71.84 & 13.71 / 70.08\\
        MCTS, (c=3, r=2) & 75.05 / 82.63 & 34.96 / 76.33 & 18.69 / 71.25 & 11.53 / 68.67\\\midrule
        MCTS, (c=4, r=1) & 79.48 / 85.92 & 39.16 / 78.11 & 20.39 / 72.04 & 14.11 / 70.10\\
        MCTS, (c=4, r=2) & 77.76 / 84.65 & 38.25 / 77.63 & 20.00 / 71.92 & 14.26 / 69.91\\
        MCTS, (c=4, r=3) & 75.65 / 82.99 & 35.03 / 75.68 & 18.80 / 71.88 & 12.36 / 68.81\\
    \bottomrule
    \end{tabular}
    \caption{
    Evaluation results studying the effects of (chosen, rejected) response quality.
    We provide results for $k=4$, as well as for both RS- and MCTS-based data curation methods.
    Each $(c=n_1,r=n_2)$ indicates that the chosen response correctly addresses $n_1$ constraints and the rejected response correctly addresses $n_2$ constraints.
    The left score indicates the hard score (i.e., the proportion of responses that get \textit{all} constraints correct), and the right score indicates the soft score (i.e., the proportion of all constraints that are satisfied by the responses).
    }
    \label{tab:response_quality_results_k4}
\end{table}

\begin{table}[!h]
    \centering
    \scriptsize
    \begin{tabular}{lcccc}
    \toprule
        & \multicolumn{4}{c}{\textbf{training k=5}}\\
        Method & IFEval & $\text{Ours}_{k=4}$ & $\text{Ours}_{k=5}$ & $\text{Ours}_{k=6}$\\\midrule
        RS, (c=4, r=1) & 76.56 / 83.74 & 36.08 / 76.81 & 19.44 / 71.42 & 11.88 / 68.84 \\
        RS, (c=4, r=2) & 74.57 / 82.17 & 34.09 / 75.82 & 17.95 / 70.71 & 10.93 / 68.35 \\
        RS, (c=4, r=3) & 72.80 / 80.97 & 31.50 / 74.37 & 16.68 / 69.94 & 10.01 / 67.19 \\\midrule
        RS, (c=5, r=2) & 76.32 / 83.55 & 35.70 / 76.64 & 19.46 / 71.51 & 12.61 / 69.39 \\
        RS, (c=5, r=3) & 74.13 / 81.92 & 33.30 / 75.35 & 17.31 / 70.15 & 11.21 / 68.08 \\
        RS, (c=5, r=4) & 72.40 / 80.42 & 30.72 / 73.75 & 16.26 / 69.31 & 9.51 / 66.76 \\\midrule
        MCTS, (c=4, r=1) & 76.71 / 83.90 & 37.24 / 77.23 & 19.49 / 71.52 & 12.37 / 69.19\\
        MCTS, (c=4, r=2) & 75.23 / 82.86 & 35.65 / 76.47 & 18.74 / 71.32 & 12.17 / 69.06 \\
        MCTS, (c=4, r=3) & 74.23 / 82.16 & 32.39 / 74.63 & 17.22 / 69.78 & 9.73 / 67.25 \\\midrule
        MCTS, (c=5, r=2) & 76.59 / 83.81 & 37.81 / 77.58 & 20.06 / 72.01 & 12.93 / 69.85 \\
        MCTS, (c=5, r=3) & 76.14 / 83.50 & 36.04 / 76.61 & 19.39 / 71.49 & 12.37 / 69.32 \\
        MCTS, (c=5, r=4) & 74.06 / 81.84 & 32.14 / 74.62 & 17.81 / 70.37 & 10.52 / 67.89 \\
    \bottomrule
    \end{tabular}
    \caption{
    Evaluation results studying the effects of (chosen, rejected) response quality.
    We provide results for $k=5$, as well as for both RS- and MCTS-based data curation methods.
    Each $(c=n_1,r=n_2)$ indicates that the chosen response correctly addresses $n_1$ constraints and the rejected response correctly addresses $n_2$ constraints.
    The left score indicates the hard score (i.e., the proportion of responses that get \textit{all} constraints correct), and the right score indicates the soft score (i.e., the proportion of all constraints that are satisfied by the responses).
    }
    \label{tab:response_quality_results_k5}
\end{table}

\begin{table}[!h]
    \centering
    \scriptsize
    \begin{tabular}{lcccc}
    \toprule
        & \multicolumn{4}{c}{\textbf{training k=4}}\\
        Method & IFEval & $\text{Ours}_{k=4}$ & $\text{Ours}_{k=5}$ & $\text{Ours}_{k=6}$\\\midrule
        MCTS, (c=4, r=1) & 78.49 / 85.36 & 39.03 / 77.76 & 22.19 / 72.10 & 14.89 / 69.76 \\
        MCTS, (c=4, r=(1,2)) & 79.77 / 86.31 & 38.97 / 78.03 & 21.99 / 71.91 & 15.24 / 70.26 \\
        MCTS, (c=4, r=(1,2,3)) & 79.97 / 86.51 & 39.31 / 78.22 & 22.19 / 72.14 & 15.89 / 70.36 \\\midrule
        MCTS, (c=(3,4), r=(0,1)) & 78.60 / 85.31 & 39.01 / 77.06 & 22.90 / 71.89 & 14.01 / 69.06\\
        MCTS, (c=(3,4), r=(0,1,2)) & 79.60 / 86.05 & 39.89 / 78.71 & 22.69 / 72.70 & 16.10 / 70.60 \\
        MCTS, (c=(3,4), r=(0,1,2,3)) & 79.42 / 85.84 & 39.37 / 78.38 & 22.24 / 72.48 & 15.20 / 70.26 \\
    \bottomrule
    \end{tabular}
    \caption{
    Evaluation results studying the effects of mixing preference pairs with different margins between the (chosen, rejected) responses.
    We provide results for $k=4$ with MCTS-based data curation methods.
    Each $(c=n_0,r=(n_1, n_2, n_3))$ indicates that the chosen response correctly addresses $n_0$ constraints and the rejected response correctly addresses either $n_1$, $n_2$ or $n_3$ constraints.
    The left score indicates the hard score (i.e., the proportion of responses that get \textit{all} constraints correct), and the right score indicates the soft score (i.e., the proportion of all constraints that are satisfied by the responses).
    }
    \label{tab:response_quality_results_mix_k4}
\end{table}

\begin{table}[!h]
    \centering
    \scriptsize
    \begin{tabular}{lcccc}
    \toprule
        & \multicolumn{4}{c}{\textbf{training k=5}}\\
        Method & IFEval & $\text{Ours}_{k=4}$ & $\text{Ours}_{k=5}$ & $\text{Ours}_{k=6}$\\\midrule
        MCTS, (c=4, r=1) & 78.88 / 85.69 & 39.55 / 78.35 & 21.88 / 71.98 & 14.59 / 69.56 \\
        MCTS, (c=4, r=(1,2)) & 78.50 / 85.30 & 37.67 / 77.55 & 21.29 / 71.51 & 14.39 / 69.15 \\
        MCTS, (c=4, r=(1,2,3)) & 78.47 / 85.25 & 39.03 / 78.06 & 22.63 / 72.49 & 15.25 / 70.05 \\\midrule
        MCTS, (c=(4,5), r=(1,2)) & 78.82 / 85.60 & 39.28 / 78.37 & 22.31 / 72.30 & 14.50 / 69.71 \\
        MCTS, (c=(4,5), r=(1,2,3)) & 78.99 / 85.57 & 38.74 / 77.89 & 21.85 / 71.87 & 13.89 / 69.00 \\
        MCTS, (c=(4,5), r=(1,2,3,4)) & 77.89 / 84.82 & 38.17 / 77.81 & 22.00 / 72.19 & 14.38 / 69.63 \\
    \bottomrule
    \end{tabular}
    \caption{
    Evaluation results studying the effects of mixing preference pairs with different margins between the (chosen, rejected) responses.
    We provide results for $k=5$ with MCTS-based data curation methods.
    Each $(c=n_0,r=(n_1, n_2, n_3))$ indicates that the chosen response correctly addresses $n_0$ constraints and the rejected response correctly addresses either $n_1$, $n_2$ or $n_3$ constraints.
    The left score indicates the hard score (i.e., the proportion of responses that get \textit{all} constraints correct), and the right score indicates the soft score (i.e., the proportion of all constraints that are satisfied by the responses).
    }
    \label{tab:response_quality_results_mix_k5}
\end{table}

\newpage
\subsection{Prompt Difficulty Results}
We provide the full results of our experiments investigating the effects of varying the difficulty of the training prompts, as measured by the number of verifiable constraints, in Tables~\ref{tab:prompt_difficulty_full_results_rs} and~\ref{tab:prompt_difficulty_full_results_mcts}.

\setlength{\tabcolsep}{8pt}
\begin{table}
    \centering
    \scriptsize
    \begin{tabular}{lcccc}
    \toprule
        & \multicolumn{4}{c}{\textbf{Rejection Sampling (RS)}}\\
        Method & IFEval & $\text{Ours}_{k=4}$ & $\text{Ours}_{k=5}$ & $\text{Ours}_{k=6}$\\\midrule
        $k=4$, (c=3, r=0) & 79.70 / 86.13 & 38.41 / 77.93 & 21.15 / 72.36 & 14.27 / 69.90\\
        $k=5$, (c=4, r=1) & 79.51 / 86.02 & 38.66 / 78.17 & 21.48 / 72.23 & 14.70 / 70.04\\
        $k=6$, (c=5, r=2) & 77.83 / 84.86 & 36.42 / 77.23 & 20.15 / 71.18 & 12.28 / 68.34\\\midrule
        $k=4$, (c=3, r=1) & 79.36 / 85.95 & 39.22 / 78.51 & 22.00 / 72.54 & 14.78 / 70.07\\
        $k=5$, (c=4, r=2) & 78.58 / 85.46 & 37.70 / 77.60 & 21.42 / 71.84 & 13.61 / 69.14\\
        $k=6$, (c=5, r=3) & 76.73 / 83.89 & 36.16 / 76.62 & 19.96 / 70.97 & 12.87 / 68.15\\
    \bottomrule
    \end{tabular}
    \caption{
    Evaluation results investigating the effects of training prompt difficulty ($k\in \{4, 5, 6\}$) on downstream performance for evaluation sets of varying difficulties.
    We provide results for RS-based preference data curation, as well as for different margins between the (chosen, rejected) responses.
    Each $(c=n_1,r=n_2)$ indicates that the chosen response correctly addresses $n_1$ constraints and the rejected response correctly addresses $n_2$ constraints.
    The left score indicates the hard score (i.e., the proportion of responses that get \textit{all} constraints correct), and the right score indicates the soft score (i.e., the proportion of all constraints that are satisfied by the responses).
    }
    \label{tab:prompt_difficulty_full_results_rs}
\end{table}

\begin{table}
    \centering
    \scriptsize
    \begin{tabular}{lcccc}
    \toprule
        & \multicolumn{4}{c}{\textbf{Monte Carlo Tree Search (MCTS)}}\\
        Method & IFEval & $\text{Ours}_{k=4}$ & $\text{Ours}_{k=5}$ & $\text{Ours}_{k=6}$\\\midrule
        $k=4$, (c=3, r=0) & 78.94 / 85.62 & 39.00 / 78.10 & 21.60 / 72.54 & 14.40 / 70.04 \\
        $k=5$, (c=4, r=1) & 78.72 / 85.40 & 39.42 / 78.36 & 21.43 / 72.27 & 14.25 / 69.95 \\
        $k=6$, (c=5, r=2) & 78.13 / 85.05 & 38.55 / 77.88 & 21.64 / 72.23 & 14.45 / 69.81 \\\midrule
        $k=4$, (c=3, r=1) & 79.12 / 85.83 & 38.81 / 78.24 & 21.72 / 72.48 & 15.66 / 70.45 \\
        $k=5$, (c=4, r=2) & 77.53 / 84.65 & 38.93 / 78.16 & 21.69 / 72.41 & 15.20 / 70.62 \\
        $k=6$, (c=5, r=3) & 77.18 / 84.53 & 37.06 / 77.10 & 20.16 / 71.25 & 12.84 / 68.57 \\
    \bottomrule
    \end{tabular}
    \caption{
    Evaluation results investigating the effects of training prompt difficulty ($k\in \{4, 5, 6\}$) on downstream performance for evaluation sets of varying difficulties.
    We provide results for MCTS-based preference data curation, as well as for different margins between the (chosen, rejected) responses.
    Each $(c=n_1,r=n_2)$ indicates that the chosen response correctly addresses $n_1$ constraints and the rejected response correctly addresses $n_2$ constraints.
    The left score indicates the hard score (i.e., the proportion of responses that get \textit{all} constraints correct), and the right score indicates the soft score (i.e., the proportion of all constraints that are satisfied by the responses).
    }
    \label{tab:prompt_difficulty_full_results_mcts}
\end{table}

\subsection{SFT vs. DPO}
We provide the full results of our additional experiments comparing the performances of models trained via SFT and DPO in Tables~\ref{tab:sft_dpo_full_results_rs} and~\ref{tab:sft_dpo_full_results_mcts}.

\setlength{\tabcolsep}{8.0pt}
\begin{table}
    \centering
    \scriptsize
    \begin{tabular}{lcccc}
    \toprule
        & \multicolumn{4}{c}{\textbf{Rejection Sampling (RS)}}\\
        Method & IFEval & $\text{Ours}_{k=4}$ & $\text{Ours}_{k=5}$ & $\text{Ours}_{k=6}$\\\midrule
        policy (\texttt{llama-3.1-8b-instruct}) & 71.71 / 80.12 & 27.77 / 71.53 & 14.15 / 67.76 & 7.34 / 64.43 \\\midrule
        SFT, $k=4$, (c=4, r=1/2/3) & 74.77 / 82.58 & 28.17 / 72.40 & 15.10 / 68.04 & 7.10 / 64.36 \\
        DPO, $k=4$, (c=4, r=1/2/3) & 78.70 / 85.48 & 36.68 / 76.78 & 20.20 / 70.71 & 13.57 / 68.45 \\\midrule
        SFT, $k=5$, (c=4, r=1/2/3) & 74.75 / 82.59 & 28.36 / 72.42 & 14.53 / 68.37 & 7.74 / 65.07 \\
        DPO, $k=5$, (c=4, r=1/2/3) & 79.08 / 85.56 & 37.98 / 77.83 & 20.50 / 71.48 & 13.42 / 68.86 \\
    \bottomrule
    \end{tabular}
    \caption{
    Evaluation results comparing the performance of training on our preference datasets via DPO compared to the base policy model or running SFT on the chosen responses only.
    We provide results for RS-based preference data curation, as well as for different training prompt difficulties ($k=4$ or $k=5$).
    Each $(c=n_1,r=n_2)$ indicates that the chosen response correctly addresses $n_1$ constraints and the rejected response correctly addresses $n_2$ constraints.
    The left score indicates the hard score (i.e., the proportion of responses that get \textit{all} constraints correct), and the right score indicates the soft score (i.e., the proportion of all constraints that are satisfied by the responses).
    }
    \label{tab:sft_dpo_full_results_rs}
\end{table}

\begin{table}
    \centering
    \scriptsize
    \begin{tabular}{lcccc}
    \toprule
        & \multicolumn{4}{c}{\textbf{Monte Carlo Tree Search (MCTS)}}\\
        Method & IFEval & $\text{Ours}_{k=4}$ & $\text{Ours}_{k=5}$ & $\text{Ours}_{k=6}$\\\midrule
        policy (\texttt{llama-3.1-8b-instruct}) & 71.71 / 80.12 & 27.77 / 71.53 & 14.15 / 67.76 & 7.34 / 64.43 \\\midrule
        SFT, $k=4$, (c=4, r=1/2/3) & 73.30 / 81.83 & 29.70 / 73.48 & 15.02 / 69.16 & 7.50 / 65.56 \\
        DPO, $k=4$, (c=4, r=1/2/3) & 79.97 / 86.51 & 39.31 / 78.22 & 22.19 / 72.14 & 15.89 / 70.36 \\\midrule
        SFT, $k=5$, (c=4, r=1/2/3) & 73.87 / 81.89 & 28.54 / 72.37 & 13.49 / 67.54 & 6.68 / 64.13 \\
        DPO, $k=5$, (c=4, r=1/2/3) & 78.47 / 85.25 & 39.03 / 78.06 & 22.63 / 72.49 & 15.25 / 70.05 \\
    \bottomrule
    \end{tabular}
    \caption{
    Evaluation results comparing the performance of training on our preference datasets via DPO compared to the base policy model or running SFT on the chosen responses only.
    We provide results for MCTS-based preference data curation, as well as for different training prompt difficulties ($k=4$ or $k=5$).
    Each $(c=n_1,r=n_2)$ indicates that the chosen response correctly addresses $n_1$ constraints and the rejected response correctly addresses $n_2$ constraints.
    The left score indicates the hard score (i.e., the proportion of responses that get \textit{all} constraints correct), and the right score indicates the soft score (i.e., the proportion of all constraints that are satisfied by the responses).
    }
    \label{tab:sft_dpo_full_results_mcts}
\end{table}

\end{document}